\documentclass{article}

\usepackage{microtype}
\usepackage{graphicx}
\usepackage{enumitem}

\usepackage{subfigure}
\usepackage{amsmath,amsfonts,amsthm,amssymb,mathtools}
\usepackage{booktabs} % for professional tables
\usepackage{multirow}
\usepackage{tikz}
\usetikzlibrary{shapes.geometric, arrows}
\usetikzlibrary{decorations.pathreplacing}
\usetikzlibrary{fit}
\usetikzlibrary{calc}
\usetikzlibrary{backgrounds}
\usetikzlibrary{patterns}
\usetikzlibrary{positioning,arrows.meta,bending}
\tikzstyle{layer} = [rectangle, text centered, draw=black, fill=red!30, style={inner sep=0,outer sep=0}, minimum height=0.25cm]
\tikzstyle{augmented}=[rectangle, draw=black, pattern=north east lines, pattern color=black]
\tikzstyle{arrow} = [thick,->,>=stealth]
\tikzstyle{layertext} = [text centered]
\tikzstyle{background}=[rectangle, draw=black, dashed, style={inner sep=0.2cm}]
\tikzstyle{backgroundlegend}=[rectangle, draw=black, dashed, style={inner sep=0.15cm}]
\tikzstyle{layerlegend} = [rectangle, draw=black, fill=red!30, style={inner sep=0.15cm}]
\tikzstyle{auglegend} = [rectangle, draw=black, pattern=north east lines, pattern color=black, style={inner sep=0.15cm}]

\usepackage{hyperref}

\usepackage[accepted]{icml2020}
\usepackage{color}
\usepackage{bm}
\usepackage{hyperref}
\usepackage{amsmath,amsfonts}
\usepackage{blindtext}
\usepackage{listings}
\usepackage{mathtools}

\newif\ifdebug
\debugtrue
\DeclarePairedDelimiterX{\infdivx}[2]{(}{)}{%
	#1\;\delimsize\|\;#2%
}

\newcommand{\vect}[1]{\boldsymbol{\mathbf{#1}}}

\newcommand{\E}{\mathbb{E}}
\newcommand{\R}{\mathbb{R}}

\newcommand{\epsilonv}{\vect\epsilon}

\newcommand{\thetav}{\vect\theta}

\newcommand{\muv}{\vect\mu}

\newcommand{\piv}{\vect\pi}

\newcommand{\sigmav}{\vect\sigma}

\newcommand{\phiv}{\vect\phi}

\newcommand{\Thetav}{\vect\Theta}

\newcommand{\Phiv}{\vect\Phi}

\newcommand{\av}{\vect a}
\newcommand{\bv}{\vect b}

\newcommand{\fv}{\vect f}
\newcommand{\gv}{\vect g}
\newcommand{\hv}{\vect h}

\newcommand{\mv}{\vect m}
\newcommand{\nv}{\vect n}

\newcommand{\sv}{\vect s}

\newcommand{\uv}{\vect u}
\newcommand{\wv}{\vect w}
\newcommand{\xv}{\vect x}
\newcommand{\yv}{\vect y}
\newcommand{\zv}{\vect z}

\newcommand{\Iv}{\vect I}

\newcommand{\Wv}{\vect W}

\newcommand{\Nc}{\mathcal N}

\newcommand{\Pc}{\mathcal P}
\newcommand{\Qc}{\mathcal Q}

\newcommand{\Rb}{\mathbb R}

\newcommand{\abs}[1]{\left|#1\right|}

\newcommand{\splitop}{\mbox{split}}
\newcommand{\concat}{\mbox{concat}}

\newtheorem{theorem}{Theorem}
\newtheorem{corollary}{Corollary}
\icmltitlerunning{VFlow: More Expressive Generative Flows with  Variational Data Augmentation}

\begin{document}

\twocolumn[
\icmltitle{VFlow: More Expressive Generative Flows with \\ Variational Data Augmentation}




\begin{icmlauthorlist}
\icmlauthor{Jianfei Chen}{thu,realai}
\icmlauthor{Cheng Lu}{thu}
\icmlauthor{Biqi Chenli}{thu,realai}
\icmlauthor{Jun Zhu}{thu,realai}
\icmlauthor{Tian Tian}{thu,realai}
\end{icmlauthorlist}

\icmlaffiliation{thu}{Department of Computer Science and Technology, Institute for
AI, BNRist Center, Tsinghua University}
\icmlaffiliation{realai}{RealAI}

\icmlcorrespondingauthor{Jun Zhu}{dcszj@mail.tsinghua.edu.cn}
%\icmlcorrespondingauthor{Eee Pppp}{ep@eden.co.uk}

\icmlkeywords{Machine Learning, ICML}

\vskip 0.3in
]



\printAffiliationsAndNotice{}  % leave blank if no need to mention equal contribution

\begin{abstract}
\frenchspacing
Generative flows are promising tractable models for density modeling that define probabilistic distributions with invertible transformations. 
However, tractability imposes architectural constraints on generative flows. 
In this work, we study a previously overlooked constraint that all the intermediate representations must have the same dimensionality with the data due to invertibility, limiting the width of the network. 
We propose VFlow to tackle this constraint on dimensionality. VFlow augments the data with extra dimensions and defines a maximum evidence lower bound (ELBO) objective for estimating the distribution of augmented data jointly with the variational data augmentation distribution. 
Under mild assumptions, we show that the maximum ELBO solution of VFlow is always better than the original maximum likelihood solution. 
For image density modeling on the CIFAR-10 dataset, VFlow achieves a new state-of-the-art 2.98 bits per dimension.
\end{abstract}

\section{Introduction}

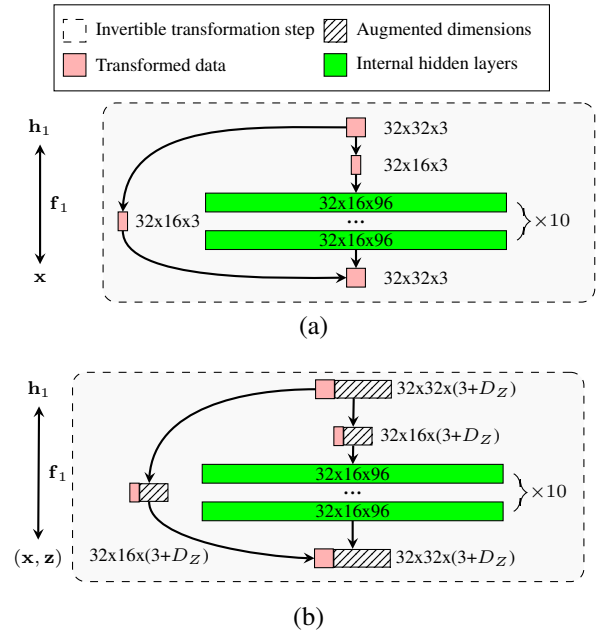
\begin{figure}[t]
	%\centering
	\hspace{0.2cm}
	\begin{tikzpicture}[node distance=0.5cm]

\node(layer1)[layer, minimum width=0.25cm] {\scriptsize };
\node(layer1text)[layertext, xshift=0.3cm, right of=layer1] {\scriptsize 32x32x3};
\node(layer1annotation)[layertext, xshift=-3.7cm, left of=layer1] {\scriptsize $\xv$};
\node(layer2)[layer, above of=layer1, minimum width=4.0cm, fill=green] {\scriptsize 32x16x96};
\draw [arrow] (layer2) -> (layer1);
\node(layer3)[layer, above of=layer2, minimum width=4.0cm, fill=green] {\scriptsize 32x16x96};
\draw[arrow, opacity = 0] (layer3) -> node[opacity = 1.0]{...}(layer2);
\draw [decorate,decoration={brace,amplitude=5pt,mirror,raise=4pt},yshift=0pt] (layer2.east) -- (layer3.east) node [black,midway,xshift=0.6cm] (coupling1annotation) {\scriptsize $\times 10$};
\node(layer4)[layer, above of=layer3, minimum width=0.125cm] {\scriptsize };
\node(layer4text)[layertext, xshift=0.3cm, right of=layer4] {\scriptsize 32x16x3};
\draw [arrow] (layer4) -> (layer3);
\node(layer5)[layer, above of=layer4, minimum width=0.25cm] {\scriptsize };
\node(layer5text)[layertext, xshift=0.3cm, right of=layer5] {\scriptsize 32x32x3};
\node(layer5annotation)[layertext, xshift=-3.7cm, left of=layer5] {\scriptsize $\hv_1$};
\draw [arrow] (layer5) -> (layer4);
\node(layer6)[layer, left of=coupling1annotation, xshift=-5.2cm, minimum width=0.125cm] {\scriptsize };
\node(layer6text)[layertext, xshift=0.1cm, right of=layer6] {\scriptsize 32x16x3};

\draw [arrow, style=<->] (layer1annotation) -> node [right] {\scriptsize $\fv_{1}$} (layer5annotation);

\begin{pgfonlayer}{background}
    \node [background,
                fit=(layer1) (layer5) (layer6) (coupling1annotation),
                rounded corners=5, fill=gray!5] {};
\end{pgfonlayer}

\begin{scope}[every edge/.append style={thick,->,>=stealth}]
 \path (layer5.west) edge[out=180,in=90] (layer6.north);
 \path (layer6.south) edge[out=270,in=180,looseness=0.6] (layer1.west);
\end{scope}

\matrix [draw,below left,above right=of layer5, xshift=-4.5cm] {
  \node [backgroundlegend,label=right:\scriptsize{Invertible transformation step}] {}; 
  & 
  \node [auglegend,label=right:\scriptsize{Augmented dimensions}] {}; \\
  \\
  \node [layerlegend,label=right:\scriptsize{Transformed data}] {}; 
  &
  \node [layerlegend,fill=green, label=right:\scriptsize{Internal hidden layers}] {}; \\
};

\end{tikzpicture}\\
\vspace{-0.1cm}
\begin{minipage}{\linewidth}
\centering
(a)
\end{minipage}\vspace{0.5cm}\\\vspace{1em}
	\begin{tikzpicture}[node distance=0.5cm]

\node(layer1)[layer,minimum width=0.25cm] {\scriptsize };
\node(layer1aug)[augmented, right=0cm of layer1, minimum width=0.75cm] {\scriptsize };
\node(layer1text)[layertext, xshift=1.25cm, right of=layer1] {\scriptsize 32x32x(3+${D_Z}$)};
\node(layer1annotation)[layertext, xshift=-3.3cm, left of=layer1] {\scriptsize $(\xv,\zv)$};
\path (layer1.north west) -- (layer1aug.north east) coordinate[midway] (layer1mid);
\node(layer2)[layer, above of=layer1mid, minimum width=4.0cm, fill=green] {\scriptsize 32x16x96};
\draw [arrow] (layer2) -> ($(layer1.north west)!0.5!(layer1aug.north east)$);
\node(layer3)[layer, above of=layer2, minimum width=4.0cm, fill=green] {\scriptsize 32x16x96};
\draw[arrow, opacity = 0] (layer3) -> node[opacity = 1.0]{...}(layer2);
\draw [decorate,decoration={brace,amplitude=5pt,mirror,raise=4pt},yshift=0pt] (layer2.east) -- (layer3.east) node [black,midway,xshift=0.6cm] (coupling1annotation) {\scriptsize $\times 10$};
\node(layer4)[layer, above of=layer3, xshift=-0.1875cm, minimum width=0.125cm] {\scriptsize };
\node(layer4aug)[augmented, right=0cm of layer4, minimum width=0.375cm] {\scriptsize };
\node(layer4text)[layertext, xshift=0.85cm, right of=layer4] {\scriptsize 32x16x(3+${D_Z}$)};
\draw [arrow] ($(layer4.south west)!0.5!(layer4aug.south east)$) -> (layer3);
\path (layer4.north west) -- (layer4aug.north east) coordinate[midway] (layer4mid);
\node(layer5)[layer, above of=layer4mid, xshift=-0.375cm, minimum width=0.25cm] {\scriptsize };
\node(layer5aug)[augmented, right=0cm of layer5, minimum width=0.75cm] {\scriptsize };
\node(layer5text)[layertext, xshift=1.25cm, right of=layer5] {\scriptsize 32x32x(3+${D_Z}$)};
\node(layer5annotation)[layertext, xshift=-3.3cm, left of=layer5] {\scriptsize $\hv_1$};
\draw [arrow] ($(layer5.south west)!0.5!(layer5aug.south east)$) -> ($(layer4.north west)!0.5!(layer4aug.north east)$);
\node(layer6)[layer, left of=coupling1annotation, xshift=-5cm, minimum width=0.125cm] {\scriptsize };
\node(layer6aug)[augmented, right=0cm of layer6, minimum width=0.375cm] {\scriptsize };
\path (layer6.south west) -- (layer6aug.south east) coordinate[midway] (layer6mid);
\node(layer6text)[layertext, below=0.5cm of layer6mid] {\scriptsize 32x16x(3+${D_Z}$)};

\draw [arrow, style=<->] (layer1annotation) -> node [right] {\scriptsize $\fv_1$} (layer5annotation);

\begin{scope}[every edge/.append style={thick,->,>=stealth}]
 \path (layer5.west) edge[out=180,in=90] ($(layer6.north west)!0.5!(layer6aug.north east)$);
 \path ($(layer6.south west)!0.5!(layer6aug.south east)$) edge[out=270,in=180,looseness=0.6] (layer1.west);

\begin{pgfonlayer}{background}
    \node [background, style={inner sep=0.1cm},
                fit=(layer1) (layer5) (layer6text) (coupling1annotation),
                label=right:\scriptsize{}, rounded corners=5, fill=gray!5] {};
\end{pgfonlayer}

\end{scope}
\end{tikzpicture}\\
\begin{minipage}{\linewidth}
\vspace{-0.4cm}
\centering
(b)
\end{minipage}\\
\vspace{-0.6cm}
	\caption{(a) Bottleneck problem in a Flow++~\cite{ho2019flow++} for CIFAR-10. Dimensionality of the transformed data (red) limits the model capacity.  
	(b) Our solution VFlow, where $D_Z$ is the dimensionality of the augmented random variable. Only the transformation step $\fv_1$ is shown due to space constraint. 
	\label{fig:flowpp}}
\end{figure}

Generative flows~\cite{nice,realnvp,glow,ho2019flow++} are a promising class of   generative models. They define a probability distribution $p(\xv)$ by applying an invertible transformation $\xv=\fv^{-1}(\epsilonv)$ to some simple and known distribution $p(\epsilonv)$. Stacking a sequence $\fv_1\dots, \fv_L$ of deep neural networks as the transformation, generative flows can model complicated high-dimensional data. 
Comparing with generative adversarial networks (GANs)~\cite{gan} and variational autoencoders (VAEs)~\cite{vae}, generative flows are particularly attractive because their sampling process and density estimation are tractable. Due to these advantages, generative flows have been applied to a wide range of problems including image generation~\cite{glow}, speech synthesis~\cite{waveglow}, 3D point cloud generation~\cite{yang2019pointflow}, semi-supervised learning~\cite{nalisnick2019hybrid}, anomaly detection~\cite{choi2018waic}, and ray tracing~\cite{muller2019neural}. 

However, tractability comes with a cost of model expressiveness. 
To be tractable, generative flows have more architectural constraints compared with  other  non-invertible models, such as GANs and VAEs. One particular constraint is that the determinant of the Jacobian of $\fv$ must be efficient to compute. 
While previous work typically adopts transformations with diagonal~\cite{nice,realnvp,glow} or triangular Jacobian~\cite{papamakarios2017masked}, there has been lots of recent work developing transformations with free-form Jacobians, including invertible 1x1 convolution~\cite{glow}, continuous time flows~\cite{chen2018neural,grathwohl2018ffjord}, invertible residual blocks~\cite{behrmann2019invertible,chen2019residual}, and emerging convolutions~\cite{hoogeboom2019emerging}.

In this paper, we study another orthogonal architectural constraint, \emph{the bottleneck problem}. 
To be invertible, all the transformation steps $\fv_1, \dots, \fv_L$ must output the same dimensionality with the input data $\xv$, although each transformation (\emph{i.e.}, neural network) can have internal hidden layers of higher dimensionality.  This contradicts with the commonly adopted wisdom of deep learning to learn overcomplete features, \emph{i.e.}, higher dimensional features than the data. As an example, Fig.~\ref{fig:flowpp}(a) presents a state-of-the-art Flow++~\cite{ho2019flow++} architecture. Although each transformation $\fv_l$ has internal higher-dimensional hidden layers (green), its input and output (red) still lie on the lower-dimensional data space.  This makes the generative flow highly inefficient because the high-dimensional features extracted within  a transformation step cannot be reused by subsequent  steps. 

We propose VFlow as a solution to the bottleneck problem. VFlow augments the data $\xv$ by extra dimensions $\zv$, which are interpreted as latent variables. We develop a variational inference framework to learn a generative flow $p(\xv, \zv)$ in the augmented data space jointly with the augmented data distribution $q(\zv | \xv)$. We show that VFlow is a generalization of the vanilla generative flows, so the augmented dimensions always help. VFlow improves existing generative flows, and achieves a state-of-the-art 2.98 bits per dimension likelihood on the CIFAR-10 dataset. 

%achieves state-of-the-art modeling performance on toy data as well as the CIFAR-10 dataset, with 2.98 bits per dimension. 

On the efficiency side, the additional $q(\zv | \xv)$ network and higher data dimensionality of VFlow only add marginal overhead to the vanilla generative flows. Meanwhile, VFlow can be more compact, since more information can be shared between individual  transformation steps. Thus, each transformation step can be simpler by avoiding extracting high-dimensional features from scratch. We show that VFlow can be 2.6 times more compact than vanilla generative flows, while achieving similar model quality. Our code is open-sourced at \url{https://github.com/thu-ml/vflow}.

\section{Backgrounds}
In this section, we review the basics of generative flows and formally define the bottleneck problem.

\subsection{Generative Flows}\label{sec:flow}

Given a distribution of $D_X$-dimensional  data $\xv$ on the space $\R^{D_X}$, the task of generative modeling aims to learn a model distribution $p(\xv; \thetav)$ parameterized by $\thetav$ that approximates the data distribution. The model can be learned with the maximum likelihood principle 
\begin{align}\label{eqn:maximum-likelihood}
\max_{\thetav} \E_{\hat p(\xv)}[\log p(\xv; \thetav)],
\end{align}
where $\hat p(\xv)$ is the empirical data distribution.

Generative flows define a sequence of invertible transformation steps $\fv_1, \dots, \fv_L$, that transform a datum $\xv$ to some random variable $\epsilonv$,
\begin{align*}
\xv \overset{\fv_1}{\longleftrightarrow} \hv_1 
\overset{\fv_2}{\longleftrightarrow} \hv_2
\cdots
\overset{\fv_L}{\longleftrightarrow} \epsilonv,
\end{align*}
where $\epsilonv$ follows a simple factorized distribution that $p_{\epsilonv}(\epsilonv)=\prod_i p_{\epsilon}(\epsilon_i)$, such as the standard normal distribution. For notational simplicity, we define $\hv_0=\xv$ and $\hv_L=\epsilonv$.
Let $\fv$ be the composition of all the $L$ transformations, such that $\epsilonv = \fv(\xv; \thetav)$, a generative flow defines the model distribution with the change-of-variables formula
\begin{align*}
\log p(\xv; \thetav) = \log p_{\epsilonv}(\epsilonv) + \log \abs{\frac{\partial \epsilonv}{\partial \xv}},
\end{align*}
where $\log \abs{\frac{\partial \epsilonv}{\partial \xv}}$ is the log-absolute-determinant of the Jacobian of $\fv$. Samples from $p(\xv; \thetav)$ can be obtained by taking the inverse transformation from $p_{\epsilonv}$:
\begin{align*}
\epsilonv\sim p_{\epsilonv}, \quad \xv= \fv^{-1}(\epsilonv; \thetav).
\end{align*}
One popular invertible transformation is the affine coupling layer~\cite{realnvp}, where each transformation $\hv_l=\fv_l(\hv_{l-1}; \thetav)$ is defined as
\begin{align}
&\xv_1, \xv_2 = \splitop(\hv_{l-1}),\nonumber\\
&\yv_1=\xv_1,\quad \yv_2=\muv(\xv_1; \thetav)+\exp(\sv(\xv_1; \thetav))\circ \xv_2,\label{eqn:affine-coupling}\\
&\hv_l=\fv_l(\hv_{l-1}; \thetav)=\concat(\yv_1, \yv_2),\nonumber%\quad \log \abs{\frac{\partial \fv_{l}}{\partial \hv_{l-1}}}=\norm{\sv(\xv_1)}_1,\nonumber
\end{align}
where $\splitop(\cdot)$ is any operation that splits the input into two disjoint parts, $\concat(\cdot)$ is its inverse operation, and $\muv$, $\sv$ are neural networks with $B$ hidden layers and $D_H$ units per layer. 
\iffalse
Another useful flow layer is invertible 1x1 convolution~\cite{glow}, defined as
\begin{align}\label{eqn:invertible-1x1-conv}
\fv_l(\hv_{l-1}; \Wv_l)_{ij}=\Wv_l \hv_{l-1,ij},
\end{align}
where the subscript $_{ij}$ extracts the channels at position $(i, j)$ from a $H\times W\times C=D_X$ dimensional feature map.
\fi

\subsection{The Bottleneck Problem}\label{sec:bottleneck-problem}
Starting from the work on universal approximation theorems~\cite{gybenko1989approximation,mhaskar1993approximation} of  multi-layer perceptrons, it is well known that network width plays an important role on the model capacity. The impact of network width is also verified empirically by recent works such as Wide ResNet~\cite{zagoruyko2016wide} and EfficientNet~\cite{tan2019efficientnet}. Almost all existing non-invertible deep models, such as residual networks~\cite{he2016deep} and generative adversarial networks~\cite{gan} have features in a higher-dimensional space than the original data space.

However, for generative flows, all the transformed data  $\hv_0,\dots,\hv_L$ must have the same dimensionality $D_X$ with the input $\xv$ due to invertibility, as illustrated in Fig.~\ref{fig:flowpp}. This architecture is ineffective for three reasons: 
\begin{enumerate}[noitemsep, nolistsep, labelindent=0pt, leftmargin=*]
    \item Few features (green in Fig.~\ref{fig:flowpp}) extracted within each transformation step can pass through the bottleneck (red in Fig.~\ref{fig:flowpp}), so subsequent  transformation steps must extract their own features from scratch;
    \item For fixed dimensional $\hv_l$, the benefit of increasing the hidden layer size $D_H$ is limited. Unlike non-invertible deep networks, which can approximate arbitrary functions with large $D_H$, the capacity of a single transformation step is intrinsically limited by architectural constraints, even with infinite $D_H$. For example, an affine coupling layer~\cite{realnvp} can not alter all the dimensions at once, while an invertible residual block~\cite{behrmann2019invertible} has a bounded Lipschitz constant;
    \item Due to the limited capacity of a single transformation step, a sufficiently powerful generative flow needs to have \emph{many} transformation steps, which is expensive. 
\end{enumerate}

We refer to this issue as the \emph{bottleneck problem}. To reflect the impact of the bottleneck width on model capacity, we denote a generative flow with $D$-dimensional bottleneck as a \emph{$D$-dimensional flow}. Ideally, a $D_H$-dimensional flow completely eliminates the bottleneck. 

%\footnote{It might be possible to design invertible transformations that maps between low-dimensional space and low-dimensional manifolds in high-dimensional space for generative flows. We leave this as an open problem. }

\section{VFlow}

We present VFlow, a variational data augmentation framework and compare it with the vanilla generative flows.

\subsection{Variational Data Augmentation}
The bottleneck problem can be tackled by increasing the dimensionality of the original data, so that the dimensionality of the flow is also increased. To achieve this, we augment the data $\xv$ with an additional $D_Z$-dimensional random variable $\zv\in \R^{D_Z}$, and model the augmented data distribution $p(\xv, \zv; \thetav)$ with a  $(D_X+D_Z)$-dimensional flow. The new flow $p(\xv, \zv; \thetav)$ is more powerful since its dimensionality can be adjusted freely by setting $D_Z$.  The underlying invertible transformation becomes $\epsilonv=\fv(\xv, \zv; \thetav)$ where $\epsilonv\in \R^{D_X+D_Z}$. 

By modeling the augmented data distribution, the log marginal  likelihood $\log p(\xv; \thetav)=\log \int p(\xv, \zv; \thetav) d\zv$ and optimization problem (\ref{eqn:maximum-likelihood}) become intractable in general. Thus, we resort to the variational methods and establish a lower bound of the marginal likelihood with a variational distribution of the augmented data $q(\zv |\xv; \phiv)$:
\begin{align}\label{eqn:elbo}
\!\!\!\!\!\!\log p(\xv; \thetav) \ge \E_{q(\zv|\xv; \phiv)}[\log p(\xv, \zv; \thetav) - \log q(\zv | \xv; \phiv)],
\end{align}
which is known as evidence lower bound (ELBO) in variational inference literature.  VFlow optimizes the following maximum ELBO objective as a surrogate of the maximum likelihood objective Eq.~(\ref{eqn:maximum-likelihood}):
\begin{align}\label{eqn:maximum-elbo}
\max_{\thetav, \phiv} \E_{\hat p(\xv) q(\zv | \xv; \phiv)}[\log p(\xv, \zv; \thetav) - \log q(\zv | \xv; \phiv)].
\end{align}
After training, density estimation can be achieved with importance sampling
\begin{align}\label{eqn:importance-sampling}
\log p(\xv; \thetav) \approx \log\left(\frac{1}{S}\sum_{i=1}^S \frac{p(\xv, \zv_i; \thetav)}{q(\zv_i | \xv; \phiv)} \right),
\end{align}
where $\zv_1, \dots, \zv_S\sim q(\zv | \xv; \phiv)$ are the $S$ samples. 

The augmented data distribution $q(\zv | \xv; \phiv)$ is modeled with another conditional flow defined with an invertible transformation $\zv=\gv^{-1}(\epsilonv_q; \xv, \phiv)$:
%\footnote{$\gv: \epsilonv_q\mapsto \zv$ is defined in the opposite direction of $\fv: \xv\mapsto \epsilonv$ because samples from $q(\zv | \xv; \phiv)$ are required at training time. Since the inverse transformation of some models such as residual flows~\cite{chen2019residual} relys on expensive numerical procedure, it is better to utilize the forward transformation for sampling.}
$$\log q(\zv | \xv; \phiv)=\log p_{\epsilonv}(\epsilonv_q) - \log \abs{\frac{\partial \zv}{\partial \epsilonv_q}},$$
where $\epsilonv_q$ follows the same distribution $p_{\epsilonv}$ with $\epsilonv$.
Given that $\zv=\gv^{-1}(\epsilonv_q; \xv, \phiv)$ is a differentible reparameterization of $\epsilonv_q$, the ELBO in Eq.~(\ref{eqn:elbo}) can be optimized with the reparameterization trick~\cite{vae}. VFlow is illustrated in Fig. 1(b). 
By choosing different architectures for $p(\xv, \zv; \thetav)$ and $q(\zv | \xv; \phiv)$, VFlow can be combined with various existing generative flows~\cite{glow,ho2019flow++,chen2019residual}, and improve their expressiveness and efficiency. 

\subsection{Connection to Vanilla Generative Flows}\label{sec:connection-to-flows}
While VFlow tackles the bottleneck problem, it only maximizes a lower bound of the likelihood. It is thus worth studying whether the gain from increased dimensionality of the flow surpasses the gap between the marginal likelihood and the ELBO.

% :
% \begin{enumerate}[noitemsep, nolistsep, labelindent=0pt, leftmargin=*]
%     \item The network architecture, including the type of the invertible transformation steps (such as affine coupling layer or invertible residual block), number of steps, number of hidden units, etc.;
%     \item The dimensionality $D$, i.e., size of the input; 
%     \item The model parameter.
% \end{enumerate}

We now show that VFlow is indeed better even it only optimizes a lower bound. Before presenting the theoretical results, we need to clarify the parameter space of different flow models.
\begin{itemize}[noitemsep, nolistsep, labelindent=0pt, leftmargin=*]
    \item A vanilla generative flow defines $p_x(\xv; \thetav_x)$, where $\thetav_x\in \Thetav_x$, and $\Thetav_x$ is the parameter space. 
    \item For any $D_Z>0$, a VFlow defines $p_a(\xv, \zv; \thetav_a)$, where $\zv\in \Rb^{D_Z}$,  $\thetav_a\in \Thetav_a$, and $\Thetav_a$ is the parameter space.  Marginalizing $\zv$ yields $p_a(\xv; \thetav_a)$.
    \item For any $D_Z>0$, the variational distribution is $q(\zv | \xv; \phiv)$, where $\zv\in \Rb^{D_Z}$,  $\phiv\in \Phiv$, and $\Phiv$ is the parameter space. 
\end{itemize}

With these notations, the maximum likelihood solution of vanilla generative flows (Eq.~\ref{eqn:maximum-likelihood}) can be written as $\max_{\thetav_x} \E_{\hat p(\xv)}[\log p_x(\xv; \thetav_x)]$, and the maximum ELBO solution of VFlow can be written as $\max_{\thetav_{a},\phiv}\E_{\hat p(\xv)q(\zv|\xv; \phiv)} 
	[\log p_{a}(\xv, \zv; \thetav_{a}) - \log q(\zv | \xv; \phiv)]$.

Our analysis is based on the following  assumptions:
\vskip 0.25em
\noindent\textbf{A}1\quad
\emph{(high-dimensional flow can emulate low-dimensional flow)}
For all $\thetav_{x}\in\Thetav_x$ and $D_Z> 0$,
there exists $\thetav_{a}\in \Thetav_a$, such that for all $\xv$ and $\zv$, 
$$p_a(\xv, \zv; \thetav_{a})=p_x(\xv; \thetav_{x})p_{\epsilonv}(\zv).$$

\noindent\textbf{A}2\quad
\emph{(the variational family has an identity transformation)}
For all $D_Z> 0$, there exists $\phiv\in \Phiv$, such that for all $\xv$ and $\zv$,
$q(\zv | \xv; \phiv)=p_{\epsilonv}(\zv)$, where $p_{\epsilonv}(\zv)$ is the simple factorized distribution defined in Sec.~\ref{sec:flow}.

\vskip 0.25em

%In other words, any $D_X$ dimensional flow can be emulated by a $D_X+D_Z$ dimensional flow, which applies the same transformation to $\xv$ but leaves $\zv$ unchanged. 
%In other words, the flow family has an identity mapping.

Assumptions A1 and A2 can be verified for most existing invertible transformation steps~\cite{realnvp,glow,chen2019residual}. Consider the simplest case of a linear flow $\epsilonv=\xv\thetav_{x}$, where $\thetav_x\in \Thetav_x$ is an orthonormal matrix. Taking $\thetav_{a}=\begin{bmatrix}
\thetav_{x} & \mathbf{0} \\
\mathbf{0} & \Iv
\end{bmatrix}
$  yields $p_a(\xv, \zv; \thetav_{a})=p_{\epsilonv}\left(
\begin{bmatrix}
\xv & \zv
\end{bmatrix}
\begin{bmatrix}
\thetav_{x} & \mathbf{0} \\
\mathbf{0} & \Iv
\end{bmatrix}
\right)=p_{\epsilonv}(\xv\thetav_{x})p_{\epsilonv}(\zv), 
$ satisfying Assumption A1. Moreover, $q(\zv|\xv; \Iv)=p_{\epsilonv}(\zv\Iv)=p_{\epsilonv}(\zv)$, satisfying Assumption A2.
We leave the detailed verification for Glow~\cite{glow} and Residual Flow~\cite{chen2019residual} in Appendix~\ref{sec:assumptions}. 

The following theorem compares the maximum ELBO solution Eq.~(\ref{eqn:maximum-elbo}) of VFlow with the maximum likelihood solution Eq.~(\ref{eqn:maximum-likelihood}) of vanilla generative flows.

\begin{theorem}\label{thm1}
	Under Assumptions A1 and A2, for any $D_Z>0$, we have
	\small{
		\begin{align*}
	&\max_{\thetav_x\in \Thetav_x} \E_{\hat p(\xv)}[\log p_x(\xv; \thetav_x)]  \\
	\le&  \max_{\thetav_{a}\in \Thetav_a,\phiv\in \Phiv}\E_{\hat p(\xv)q(\zv|\xv; \phiv)} 
	[\log p_{a}(\xv, \zv; \thetav_{a}) - \log q(\zv | \xv; \phiv)].
	\end{align*}}
\end{theorem}

\begin{proof}
Our proof is based on a simple construction.
Given any vanilla flow model $p_x(\xv; \theta_x)$, according to Assumptions A1 and A2, for any $D_Z>0$, we can construct 
\begin{itemize}[noitemsep, nolistsep, labelindent=0pt, leftmargin=*]
    \item $\thetav(\thetav_x)\in \Thetav_a$, such that $p_a(\xv, \zv; \thetav(\thetav_x))$ is a factorized distribution $p_a(\xv, \zv; \thetav(\thetav_x))=p_x(\xv; \thetav_x) p_{\epsilonv}(\zv)$. This is a very weak model that does not utilize $\zv$ at all.
    \item $\phiv\in \Phiv$, such that the variational distribution is trivial $q(\zv | \xv; \phiv) = p_{\epsilonv}(\zv)$.
\end{itemize}
Even using these special models, we have 
$$\log p_a(\xv, \zv; \thetav(\thetav_x)) - \log q(\zv | \xv; \thetav_x) = \log p_x(\xv; \thetav_x).$$
Now, starting from
\begin{small}
\begin{align*}
	&\max_{\thetav_x\in \Thetav_{x}}\E_{\hat p(\xv)}[\log p_x(\xv; \thetav)]\nonumber\\
	=&\max_{\thetav_a\in \Thetav_{a},\phiv\in \Phiv}\E_{\hat p(\xv) p_{\epsilonv}(\zv)}[\log p_x(\xv; \thetav) + \log p_{\epsilonv}(\zv) - \log p_{\epsilonv}(\zv)]\nonumber,\\
	\end{align*}
\end{small}
considering the special $\thetav(\thetav_x)$, we have 
\begin{small}
\begin{align*} = \max_{\thetav_x\in\Thetav_x} \E_{\hat p(\xv)}[\log p_a(\xv, \zv; \thetav(\thetav_x)) - \log p_{\epsilonv}(\zv)],\end{align*}
\end{small}
allowing the parameter of $p_a$ to be chosen freely from $\Thetav_a$, not just $\thetav(\thetav_x) \subset \Thetav_a$, we have 
\begin{small}
\begin{align*}\le \max_{\thetav_a\in \Thetav_a} \E_{\hat p(\xv)}[\log p_a(\xv, \zv; \thetav_a) - \log p_{\epsilonv}(\zv)],\end{align*}
\end{small}
replacing $p_{\epsilonv}(\zv)$ by the trivial variational distribution, we have
\begin{small}
\begin{align}= \max_{\thetav_a\in \Thetav_a} \E_{\hat p(\xv)}[\log p_a(\xv, \zv; \thetav_a) - \log q(\zv|\xv; \phiv)],\label{eqn:gaussian}\end{align}
\end{small}
allowing $\phiv$ to be chosen freely from $\Phiv$, we have
\begin{small}
\begin{align*}\le \max_{\thetav_a\in \Thetav_a, \phiv\in \Phiv} \E_{\hat p(\xv)}[\log p_a(\xv, \zv; \thetav_a) - \log q(\zv|\xv; \phiv)].\end{align*}
\end{small}
\end{proof}

\noindent \textbf{Remark 1: }
Theorem 1 does not consider optimization issues, such as convergence speed. However, as we shall see in Appendix~\ref{sec:assumptions}, it is rather simple for a VFlow to mimic a vanilla generative flow by setting some parameters to zero, due to the residual structure of transformation steps. Therefore, we hypothesize that VFlow should still be better than vanilla generative flows under the same number of optimizer iterations. This is empirically verified in Fig.~\ref{fig:curve_expb}.

\noindent\textbf{Remark 2: } Variational inference-based models such as VAEs rely heavily on the quality of the variational posterior $q(\zv; \xv, \phiv)$ to work well. Unlike VAE, VFlow is better than vanilla generative flows even with a trivial variational distribution $q(\zv|\xv)=p_{\epsilonv}(\zv)$. This can be seen from Eq.~(\ref{eqn:gaussian}).

Finally, combining Theorem~\ref{thm1} with the variational bound Eq.~(\ref{eqn:elbo}), we have
\begin{corollary}
Under Assumptions A1 and A2, for any $D_Z>0$, we have
	\small{
		\begin{align*}
\max_{\thetav_x\in \Thetav_x} \E_{\hat p(\xv)}[\log p_x(\xv; \thetav_x)] \le \max_{\thetav_a\in \Thetav_a} \E_{\hat p(\xv)}[\log p_a(\xv; \thetav_a)]
	\end{align*}}
\end{corollary}

\subsection{Efficiency}\label{sec:param-efficiency}

While VFlow makes the model more expressive, its overhead is only marginal. To see this, note that  the overhead of VFlow arises from two parts: (1) the cost of computing $q(\zv | \xv; \phiv)$ and (2) the increase of the cost for computing $p(\xv, \zv; \thetav)$ due to the increase of the dimensionality. The first cost can be small by using a much smaller network for $q(\zv | \xv; \phiv)$ than $p(\xv, \zv; \thetav)$. As an extreme case, one can eliminate the cost by adopting $q(\zv | \xv; \phiv)=p_{\epsilonv}(\zv)$, according to Remark 2.
The second cost is small because most computation of $p(\xv, \zv; \thetav)$ is spent on the internal hidden layers (green layers in Fig.~\ref{fig:flowpp}), whose time complexity is only related to the hidden layer size $D_H$ instead of the flow dimensionality $D_X+D_Z$.

On the other hand, VFlow can be more compact and efficient than a vanilla generative flow to achieve similar modeling quality, as it alleviates the ineffectiveness listed in Sec.~\ref{sec:bottleneck-problem} caused by the bottleneck problem.

\subsection{Modeling Discrete Data}\label{sec:dequant}
The discussion so far is limited to continuous data $\xv$. If the data follow a discrete distribution $P(\xv)$, an additional de-quantization step is needed to convert the data from discrete to continuous. \citet{ho2019flow++} propose to bound the discrete density with a variational dequantization distribution $r(\uv | \xv)$:
\begin{align*}
\log P(\xv)\ge \E_{r(\uv | \xv)}[\log p(\xv+\uv)-\log r(\uv | \xv)],
\end{align*}
where $\uv$ is continuous and $p(\xv+\uv)$ is a generative flow for continuous data. Combining with the ELBO Eq.~(\ref{eqn:elbo}), we obtain a lower bound for discrete data
\begin{align}
\log P(\xv)\ge \E_{r(\uv | \xv), q(\zv | \xv+\uv)}[&\log p(\xv+\uv, \zv)-\log r(\uv | \xv)\nonumber\\
&-\log q(\zv | \xv+\uv)].\label{eqn:full-elbo}
\end{align}
Estimating the marginal density $\log P(\xv)$ involves similar importance sampling procedure with Eq.~(\ref{eqn:importance-sampling}), but the samples are drawn from the joint distribution $r(\uv | \xv) q(\zv | \xv+\uv)$ of dequantization noise and augmented data. 

Although both variational dequantization and VFlow introduce variational distributions, their purposes are different. Variational dequantization aims to reduce the gap between the discrete data distribution and continuous model distribution, while VFlow aims to increase the dimensionality of the flow. These approaches are orthogonal to each other.

\section{Related Works}\label{sec:related-works}

There exists a large bulk of works on developing more flexible transformation steps, such as transformations with free-form Jacobians~\cite{grathwohl2018ffjord,behrmann2019invertible,chen2019residual}, fast fourier transformation-based invertible convolutions~\cite{hoogeboom2019emerging}, flexible coupling functions~\cite{ho2019flow++,durkan2019neural,muller2019neural}, and masked convolutional layers~\cite{hoogeboom2019emerging,song2019mintnet}. VFlow is orthogonal with these approaches since it tackles a different bottleneck of dimensionality, and can be combined with these works to create better models. 

The bottleneck problem is studied for \emph{discriminative} invertible models including neural ODEs~\cite{dupont2019augmented} and i-RevNets~\cite{jacobsen2018revnet}, where zeros are padded to the input data to increase the number of dimensions. In contrast, VFlow studies the much more challenging \emph{generative} modeling problem. For generative modeling, zero padding does not work because the padded data $(\xv, \mathbf{0})$ still lies on a $D_X$-dimensional manifold, while the distribution $p_{\epsilonv}(\epsilonv)$ is defined on a $D_X+D_Z$ dimensional space. Therefore, an invertible transformation does not exist. 
Similarly, simply replicating the data does not help. Another possible solution is reducing the number of transformations $L$ to one. While this does eliminate the bottleneck problem, the capacity of a single transformation is limited, as discussed in Sec.~\ref{sec:bottleneck-problem}.

Variational autoencoders~\cite{vae} can be understood as VFlows where both  $p(\xv, \zv; \thetav)$ and $q(\zv | \xv; \phiv)$ are generative flows with a single affine coupling layer. 
Particularly, a Gaussian VAE $p(\xv, \zv; \thetav)=\Nc(\zv; \mathbf{0}, \Iv)\Nc(\xv; \muv(\zv), \exp(\sv(\zv))^2)$ is equivalent with 
\begin{align*}
&\epsilonv_Z\sim \Nc(\mathbf{0}, \Iv), \quad \epsilonv_X\sim \Nc(\mathbf{0}, \Iv),\\
&\zv = \epsilonv_Z, \quad \xv = \muv(\epsilonv_Z)+\exp(\sv(\epsilonv_Z))\circ \epsilonv_X,
\end{align*}
which shares the same form with the affine coupling layer defined in Eq.~(\ref{eqn:affine-coupling}), despite in the opposite direction. VFlows are more general than VAEs by not assuming the hierarchical structure $p(\xv, \zv)=p(\zv)p(\xv | \zv)$. Though it is possible for VAEs to implement both $p(\zv)$ and $p(\xv | \zv)$ with generative flows~\cite{morrow2020variational,chen2016variational}, the flow $p(\xv | \zv)$ is still $D_X$-dimensional, so the bottleneck problem persists. Another line of work implement $q(\zv | \xv)$ with generative flows~\cite{kingma2016improved,rezende2015variational} but leaves $p(\xv, \zv)$ unchanged. VFlow has identical $q(\zv | \xv)$ but more powerful $p(\xv, \zv)$ than these works. 
There are also a number of works combining VAEs with autoregressive models~\cite{chen2016variational,pixelvae}. However they suffer from slow sampling due to the sequential nature of autoregressive models. Finally, while a powerful $q(\zv | \xv)$ is critical for VAEs, it is less important for VFlows, since $p(\xv, \zv)$ itself is powerful even with $q(\zv | \xv)=p_{\epsilonv}(\zv)$, as discussed in Sec.~\ref{sec:connection-to-flows}. 

Augmented Normalizing Flow (ANF)~\cite{huang2020augmented} is an independent parallel work of VFlow. Both ANF and VFlow combine generative flows and variational inference. The main difference is the theoretical guarantee. We view VFlow as a general improvement of flow models, so our theory compares the modeling quality of original and augmented data. In contrast, ANF is more closely related to variational autoencoders, and it rather focuses on the universal approximation of probability distributions in the asymptotic case. 

\section{Toy Data Experiments}
\begin{figure}[t]
	\centering
	\begin{minipage}{0.49\linewidth}
		\centering
			\includegraphics[width=.8\linewidth]{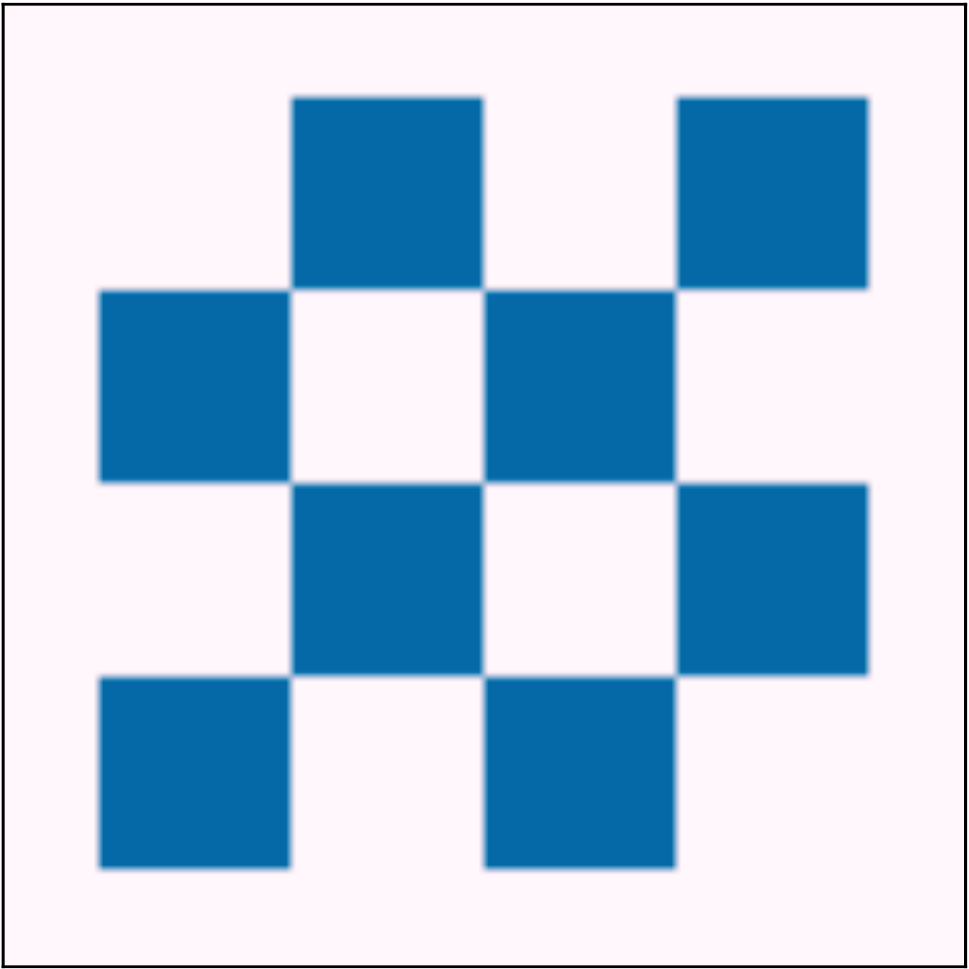}\\
\small{(a) Data (-3.47)}
	\end{minipage}
	\begin{minipage}{0.49\linewidth}
	\centering
	\includegraphics[width=.8\linewidth]{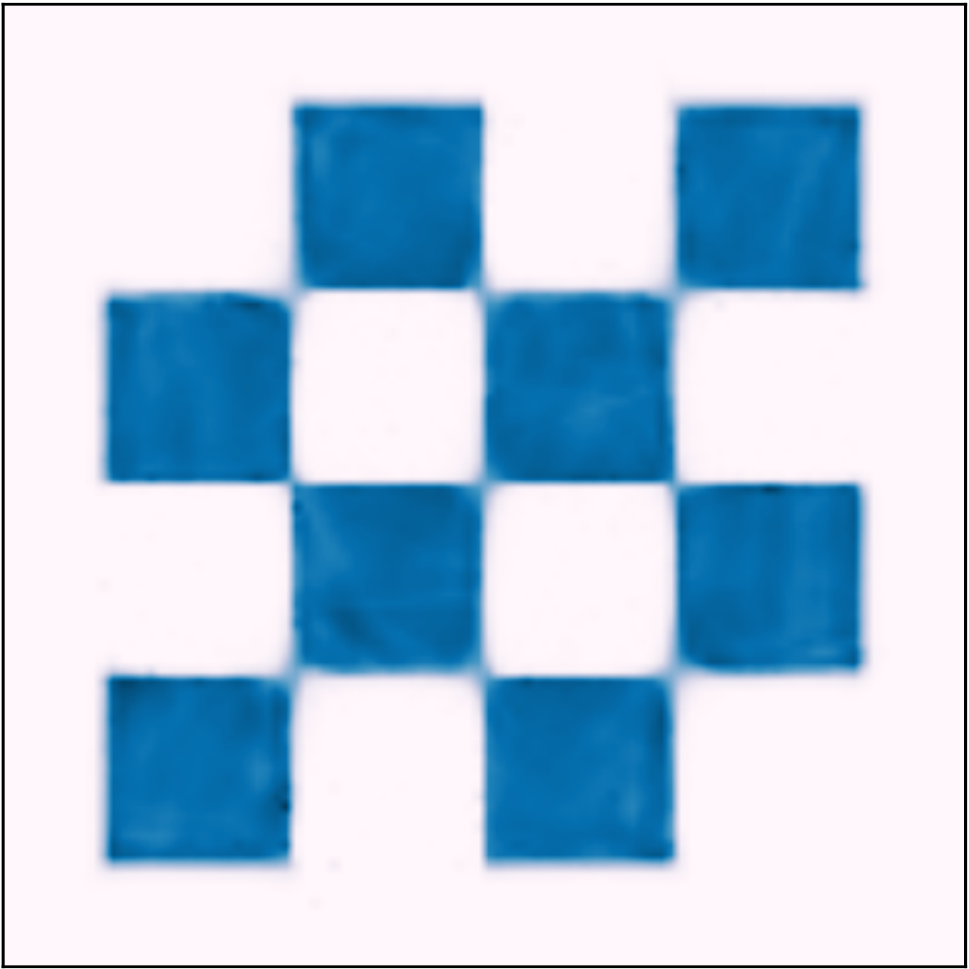}\\
\small{	(b) 3-step, 10-dim VFlow (-3.51)}
\end{minipage}\\
	\begin{minipage}{0.49\linewidth}
		\centering		
 	\includegraphics[width=.8\linewidth]{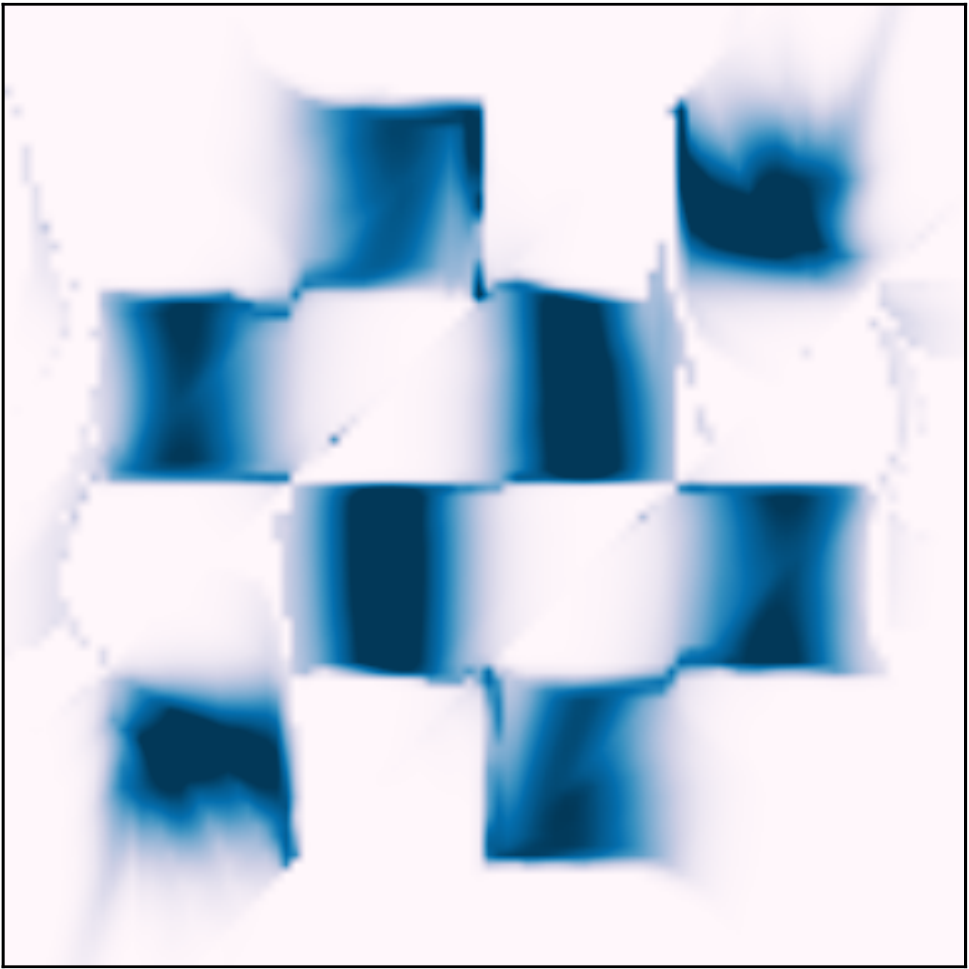}\\
\small{ 	(c) 3-step Glow (-3.66)}
   \end{minipage}
\begin{minipage}{0.49\linewidth}
	\centering		
	\includegraphics[width=.8\linewidth]{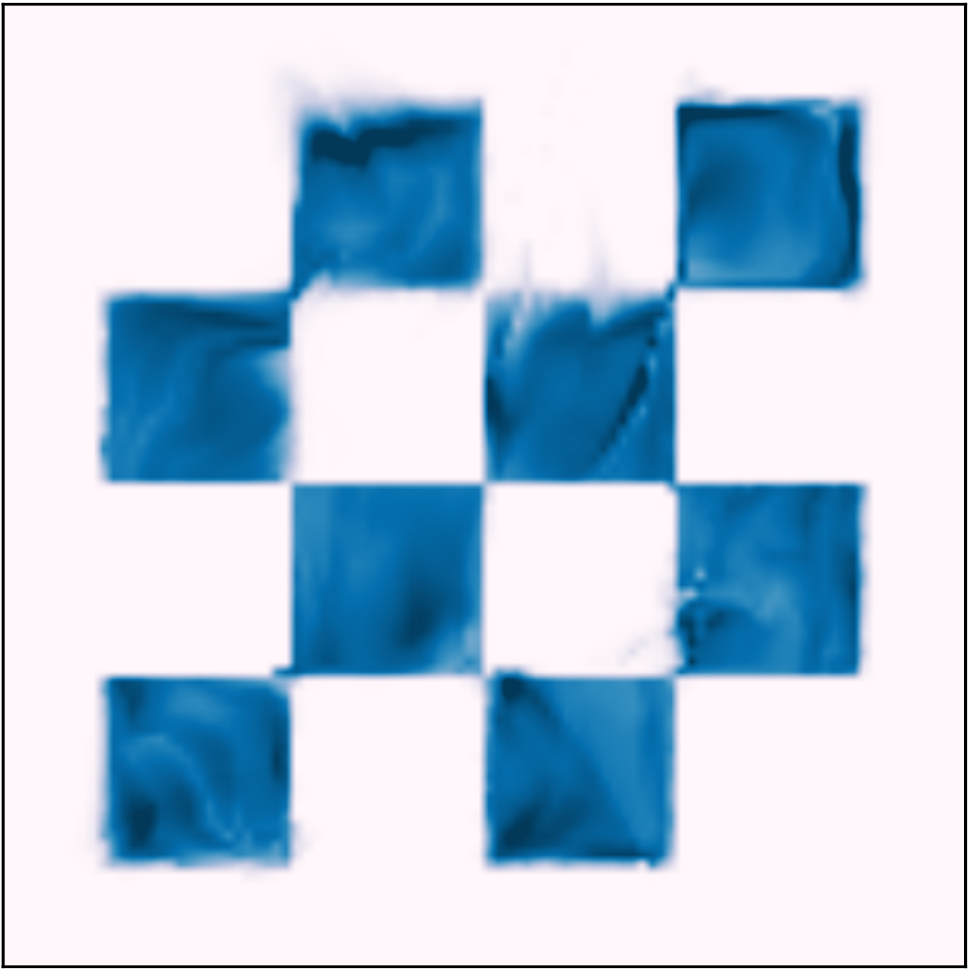}\\
	\small{(d) 20-step Glow (-3.52)}
\end{minipage}\\
	\caption{Data and model density on toy data, log-likelihood is shown in parenthesis.\label{fig:density}}
\end{figure}

\begin{figure}[t]
\centering
\includegraphics[width=.8\linewidth]{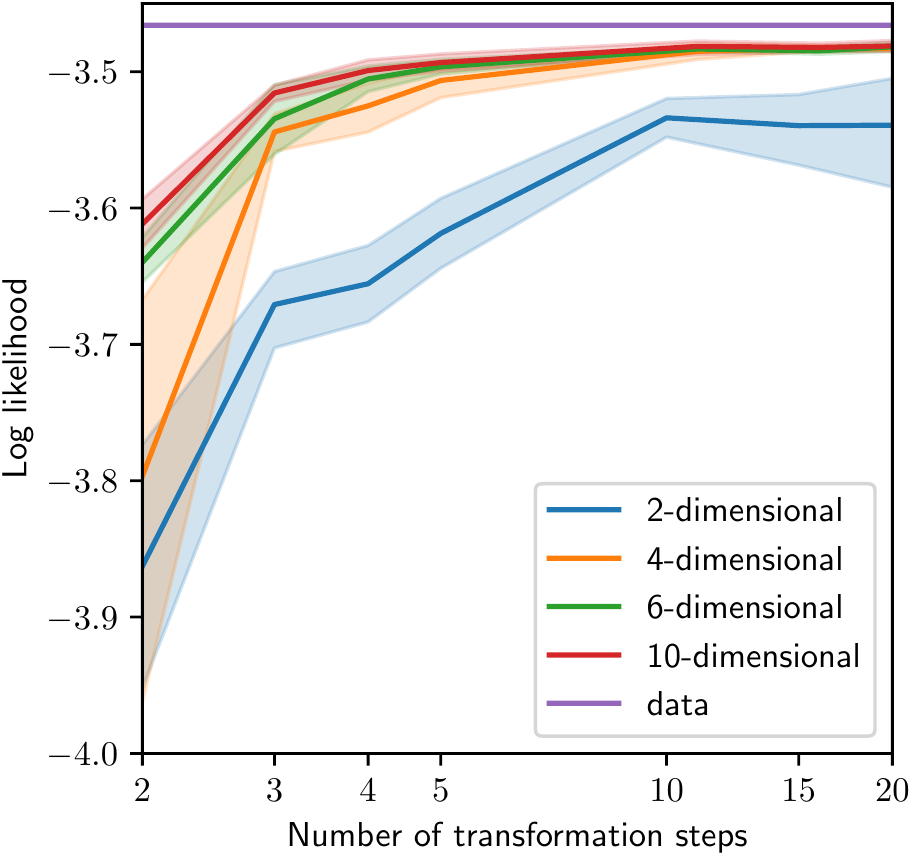}
\caption{Impact of the dimensionality  on the toy dataset.\label{fig:dims}}
\end{figure}

\begin{figure*}
\centering
\begin{tikzpicture}[node distance=4.3cm]
\node(fig1){
\begin{minipage}{0.24\linewidth}
	\centering
\includegraphics[width=.8\linewidth]{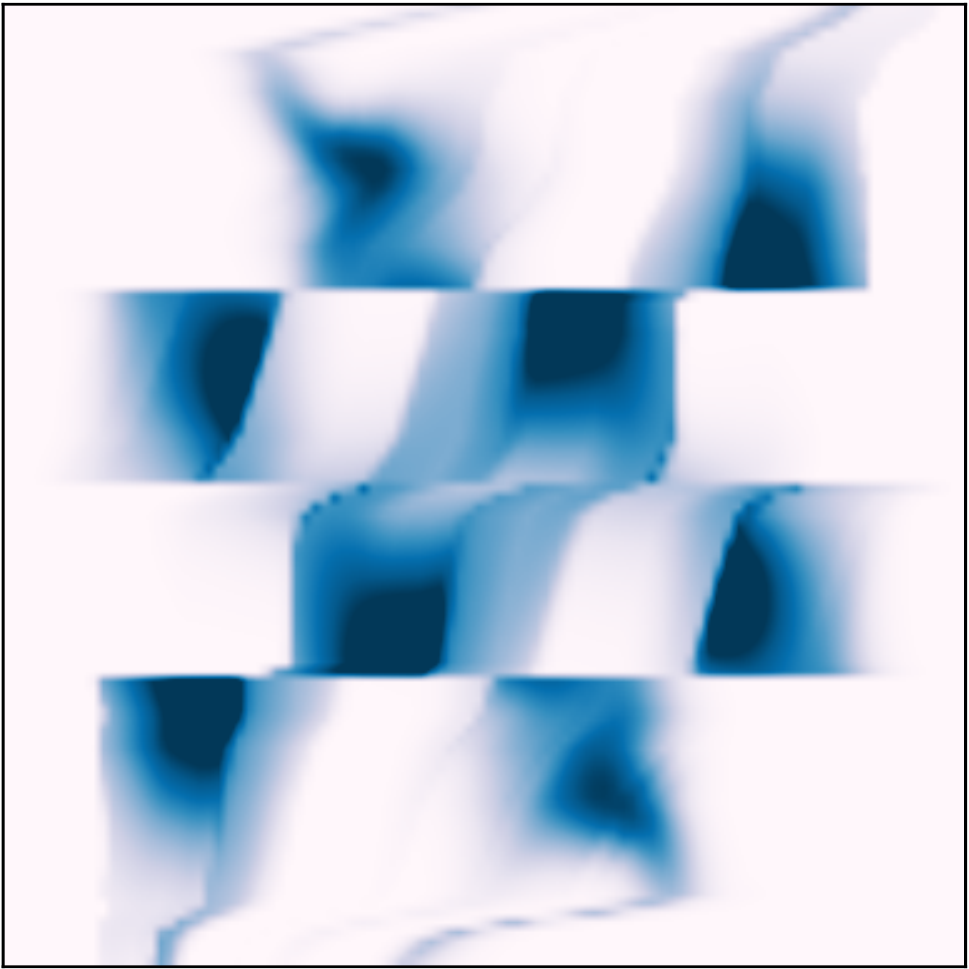}\\
Model density (-3.80)
\end{minipage}
};
\node(fig2)[right of=fig1]{
\begin{minipage}{0.24\linewidth}
		\centering
\includegraphics[width=.8\linewidth]{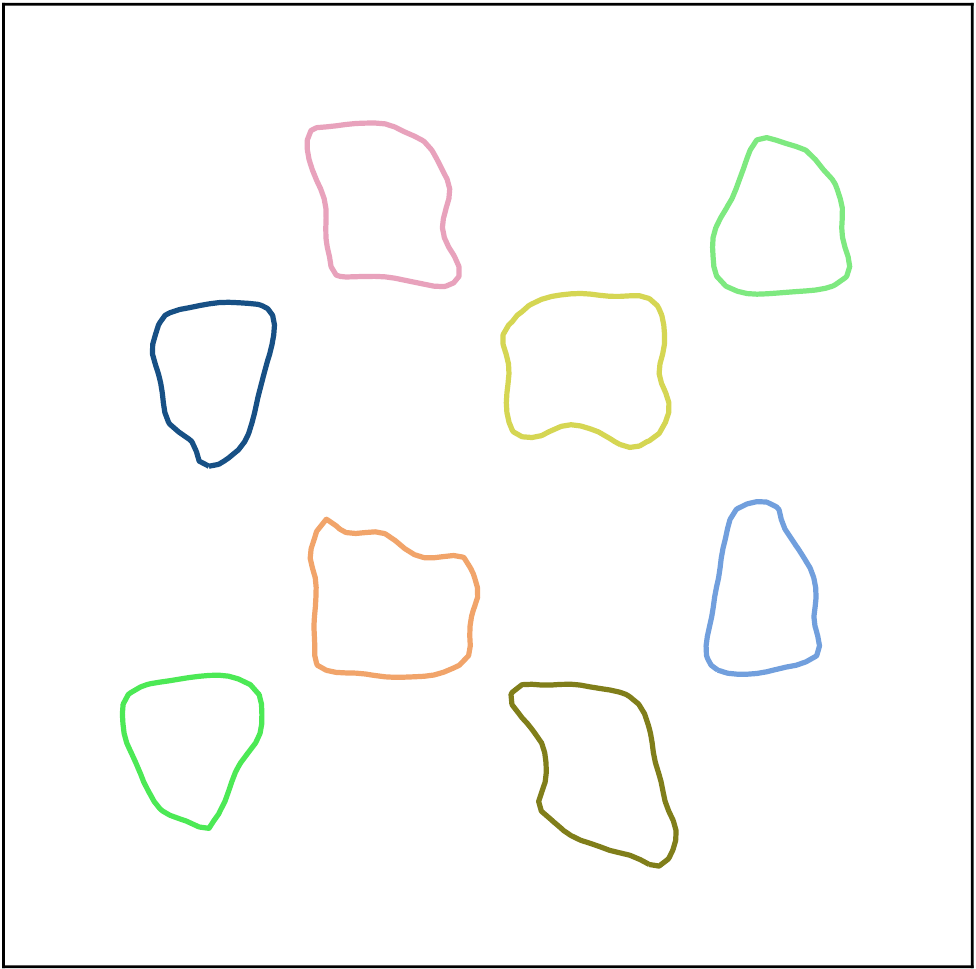}\\
$\xv$% space
\end{minipage}
};
\node(fig3)[right of=fig2]{
\begin{minipage}{0.24\linewidth}
		\centering
	\includegraphics[width=.8\linewidth]{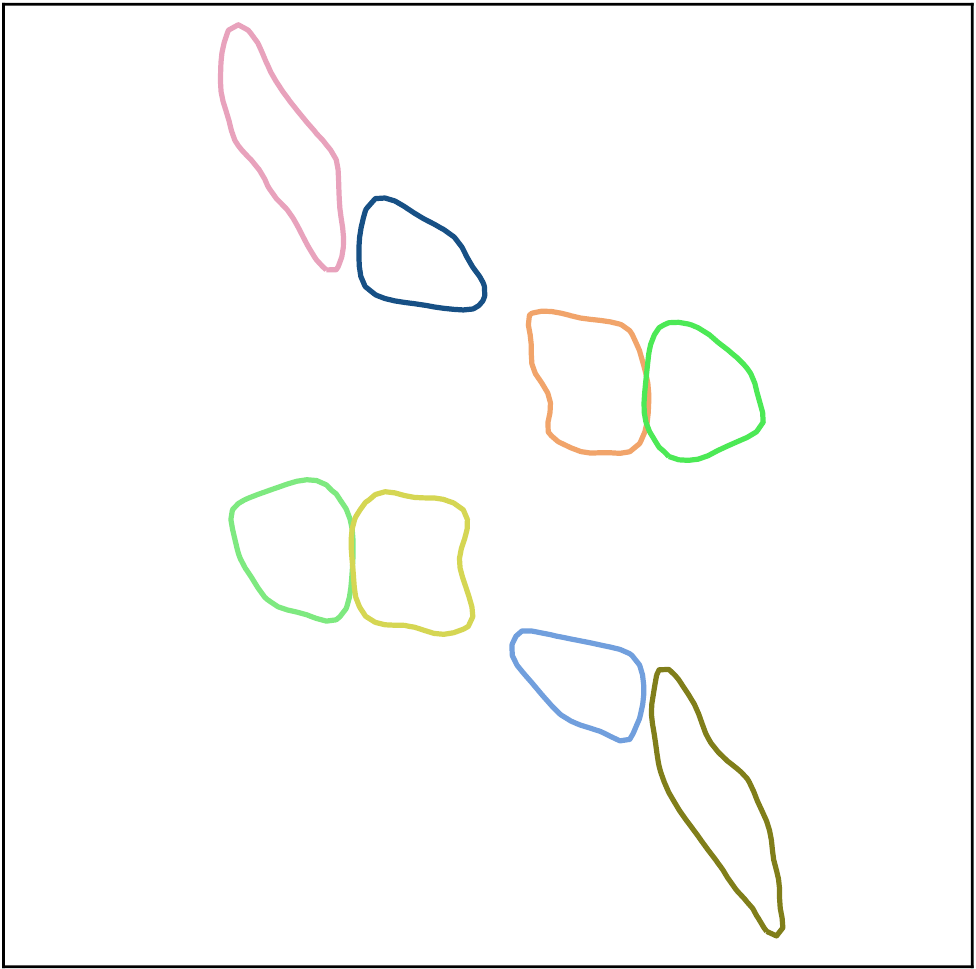}\\
	$\hv_1$% space
\end{minipage}
};
\node(fig4)[right of=fig3]{
\begin{minipage}{0.24\linewidth}
		\centering
	\includegraphics[width=.8\linewidth]{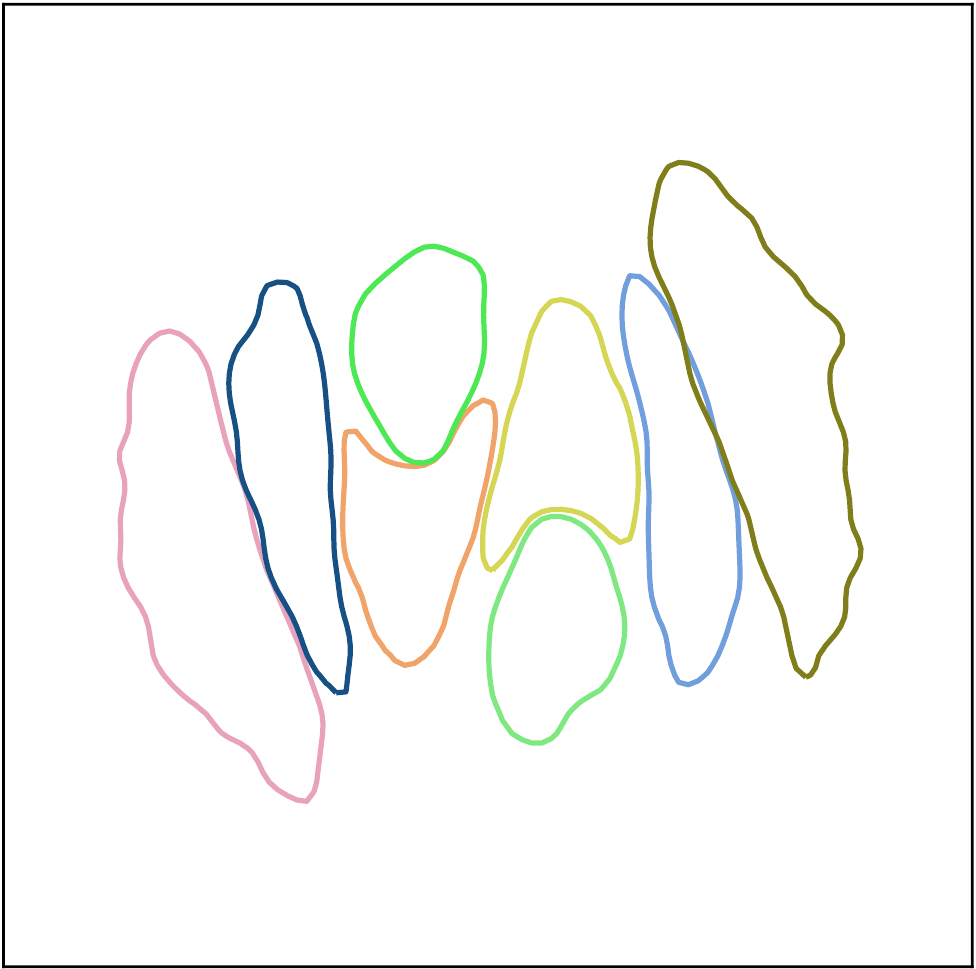} \\
	$\epsilonv$% space
\end{minipage}
};
\node(fig5)[below of=fig1]{
\begin{minipage}{0.24\linewidth}
		\centering
	\includegraphics[width=.8\linewidth]{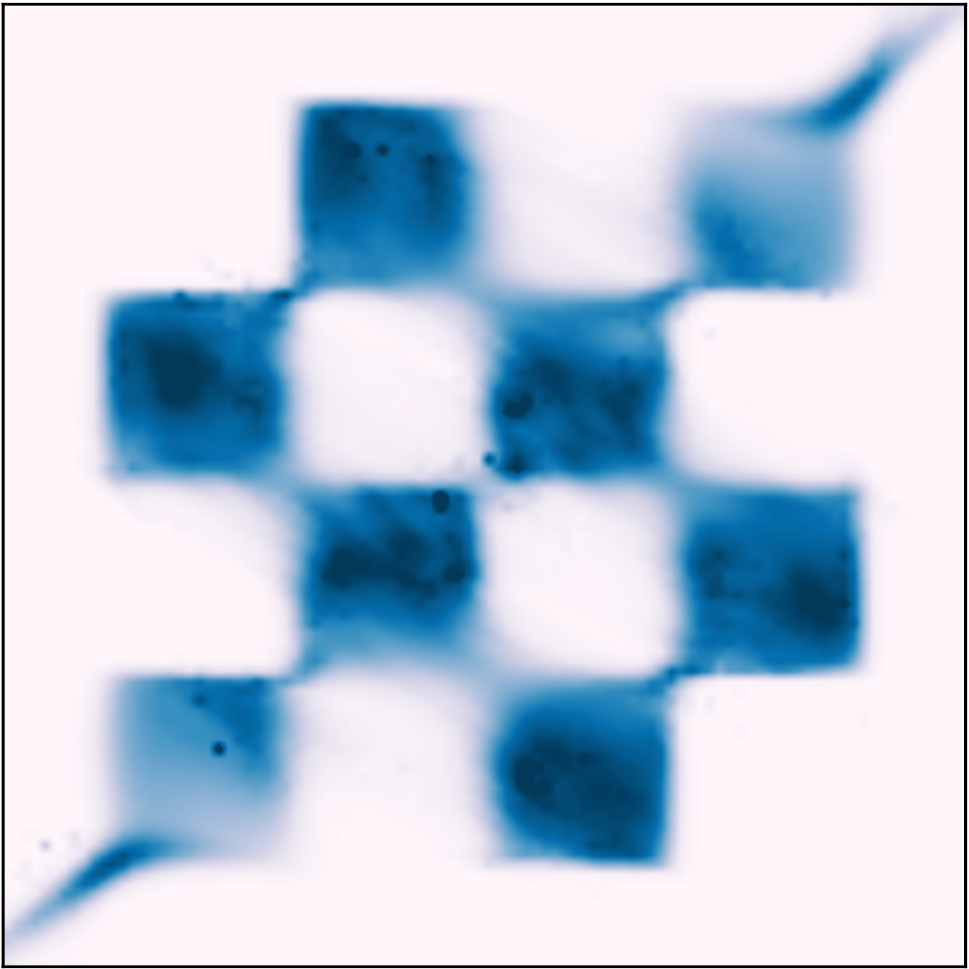}\\\vspace{1em}
Model density (-3.69)
\end{minipage}
};
\node(fig6)[right of=fig5]{
\begin{minipage}{0.24\linewidth}
		\centering
	\includegraphics[width=0.9\linewidth]{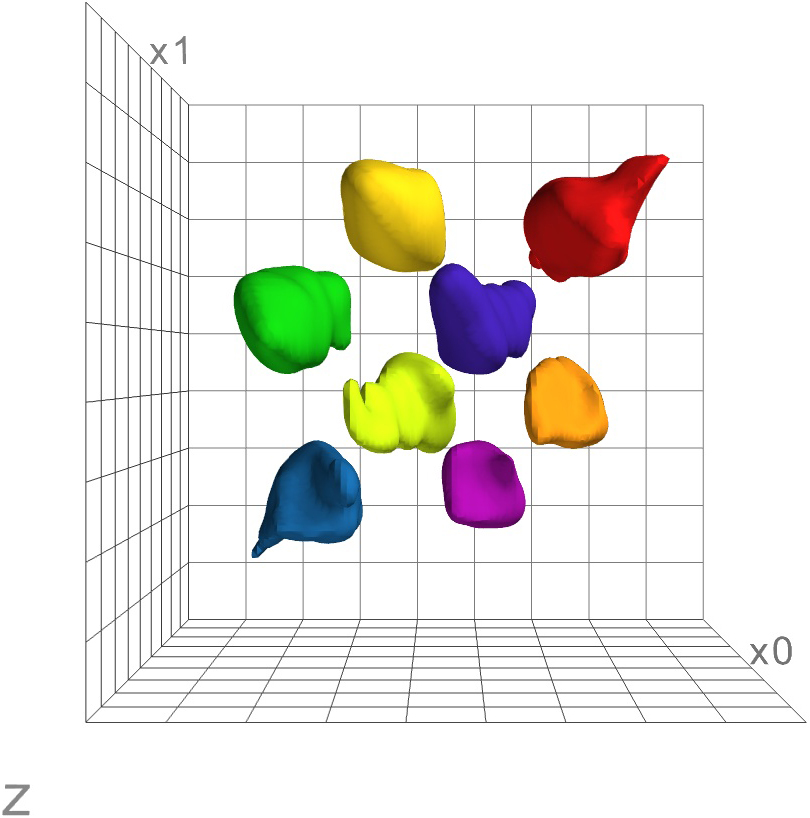}\\
	$\xv$ (top view)% space 
\end{minipage}
};
\node(fig7)[right of=fig6]{
\begin{minipage}{0.24\linewidth}
		\centering
	\includegraphics[width=0.9\linewidth]{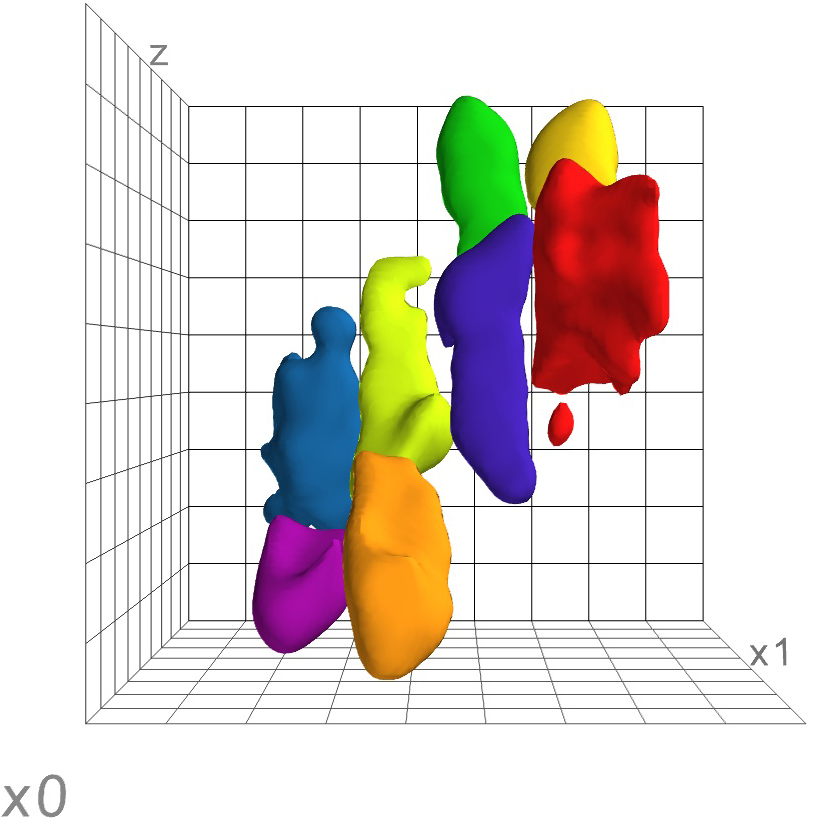}\\
	$\xv$ (front view)% space 
\end{minipage}
};
\node(fig8)[right of=fig7]{
\begin{minipage}{0.24\linewidth}
		\centering
	\includegraphics[width=0.9\linewidth]{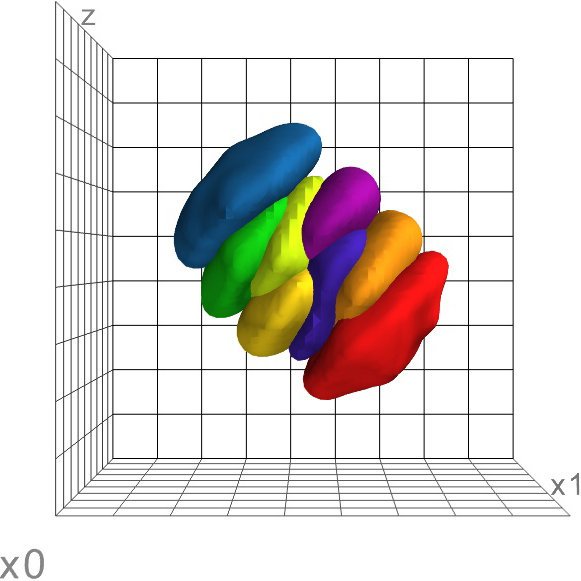}  \\
	$\epsilonv$% space
\end{minipage}
};

\draw[<->, line width=0.2mm] (6,0) -- node [above] {\scriptsize $\fv_1$} (6.9,0);
\draw[<->, line width=0.2mm] (10.3,0) -- node [above] {\scriptsize $\fv_2$} (11.2,0);
\draw[<->, line width=0.2mm] (10.7,-4) -- node [above] {\scriptsize $\fv_1$} (11.3,-4);

\draw[draw=black, dashed, rounded corners=5] (2.3,-2) rectangle ++(8.3,-4.5);

\end{tikzpicture}
\caption{Visualization of learnt transformation on toy data. Top row: 2-step Glow. Bottom row: 2-step, 3-dimensional VFlow. Log-likelihood is shown in parenthesis.  We sample $\epsilonv$ and visualize the transformed density in $\xv$, $\hv_1$ and $\epsilonv$ space. The density is estimated from samples by kernel density estimation, and we show the 50\% probability contour / isosurface for each mode in different color. 
\label{fig:vis}}
\end{figure*}

We first evaluate VFlow on a toy $D_X=2$ \emph{Checkerboard} dataset \cite{behrmann2019invertible}, which is multimodal and its density is shown in Fig.~\ref{fig:density}(a). The baseline model is Glow~\cite{glow}, where each transformation step consists of an affine coupling layer with 2 hidden layers and  $D_H=50$ hidden units per layer. VFlow further augments Glow with a conditional Glow $q(\zv | \xv; \phiv)$ and various number of extra dimensions $D_Z$. 
All the models are trained for 100,000 iterations with Adam~\cite{kingma2014adam} and a batch size of 64, and each experiment is repeated 5 times with different random seeds. Model quality is measured with the log-likelihood $\log p(\xv)$ on a 1,000-sample test set. For VFlow, likelihood is evaluated with 100-sample importance sampling by Eq.~(\ref{eqn:importance-sampling}).

We study the impact of the dimensionality of the flow $D_X+D_Z\in \{2, 4, 6, 8, 10\}$, where $D_X+D_Z=2$ is the baseline Glow and $D_X+D_Z>2$ is VFlow. To control the model size, we vary the total number of transformation steps $L\in\{2, 3, 4, 5, 10, 15, 20\}$. For baseline Glow, the $p$-network has all the $L_p=L$ transformation steps; and for VFlow, $p$-network has $L_p=L-1$ transformation steps and $q$-network has one transformation step. The result is shown in Fig.~\ref{fig:dims}, VFlow significantly outperforms Glow under similar model size. For example, a 3-step, 10-dimensional VFlow achieves $-3.51\pm 0.01$ log-likelihood (Fig.~\ref{fig:density}(b)), outperforming the baseline 3-step Glow with $-3.67\pm 0.03$ log-likelihood (Fig.~\ref{fig:density}(c)) by a large margin. The 3-layer, 10-dimensional VFlow even outperforms a much larger 20-step Glow, which achieves $-3.54\pm 0.05$ log-likelihood (Fig.~\ref{fig:density}(d)), showing that the model can be much more compact by solving the bottleneck problem.

To further understand why the dimensionality of is important, we visualize the learnt representation for a 2-step Glow and a 2-step VFlow, which has a single transformation step for both $p$ and $q$. To make visualization possible, $z$ is only one-dimensional, so $D_X+D_Z=3$. Note that having odd number of dimensions is suboptimal because the affine coupling layer cannot split the data into two parts with equal number of dimensions. 
Moreover, affine coupling layer cannot represent the one-dimensional distribution $q(z | \xv)$, so we replace it with a Gaussian layer $\Nc(z; \muv(\xv), \sigmav(\xv))$ without changing the architecture of $\muv(\xv)$ and $\sigmav(\xv)$. The learnt transformations are visualized in Fig.~\ref{fig:vis}.
While Glow struggles to map different modes to the compact space of $\epsilonv$, VFlow does a much better job. VFlow learns a pile of ``pies'' in the $\epsilonv$ space, where each mode is a pie, and different modes are directly distinguished by the extra dimension $z$.\footnote{The three axes $x_0$, $x_1$ and $z$ do not distinguish from each other in the $\epsilonv$ space  because the invertible 1x1 convolution in Glow can rotate them arbitrarily.} By shifting the pies to different positions on the $\xv$-plane based on $z$, VFlow can easily handle the multi-modality of the data. Comparing with Glow which needs to map a irregular shape in the $\epsilonv$ space to a square in the $\xv$ space, VFlow requires a much simpler transformation, because each mode is already a regular pie in the $\epsilonv$ space.

\begin{table*}[t]
    \centering
    \caption{Density modeling results in bits/dim (bpd). We report testing bpd for CIFAR-10 and validation bpd for ImageNet.}
        \vskip 0.15in
        \begin{small}
    \begin{tabular}{lccc}
    \toprule
    Model & CIFAR-10 & ImageNet\ 32x32 & ImageNet\ 64x64 \\
    \midrule
        Glow~\cite{glow} & 3.35 & 4.09 & 3.81  \\
        FFJORD~\cite{grathwohl2018ffjord} & 3.40 & \_ & \_  \\
        Residual Flow~\cite{chen2019residual} & 3.28 & 4.01 & 3.76 \\
        MintNet~\cite{song2019mintnet} & 3.32 & 4.06 & \_ \\
        Flow++~\cite{ho2019flow++} & 3.08 & 3.86 & 3.69  \\
        VFlow & \textbf{2.98} & \textbf{3.83} & \textbf{3.66} \\
        % \midrule
        % VAE-IAF~\cite{kingma2016improved} &  3.11 \\
        % Sparse Transformer~\cite{child2019sparsetransformer} &  \textbf{2.80} \\
    \bottomrule
    \end{tabular}
    \end{small}
    \label{tab:sota}
\end{table*}

\section{Density Estimation of Images}\label{sec:cifar10}
In this section, we evaluate VFlow on CIFAR-10 \footnote{\url{https://www.cs.toronto.edu/~kriz/cifar.html}} and ImageNet~\cite{ILSVRC15} for density estimation of images. VFlow augments a state-of-the-art generative flow, Flow++~\cite{ho2019flow++} by introducing extra dimensions and another variational distribution $q(\zv | \xv)$. More specifically, the $p(\xv, \zv)$ network is similar with Flow++ shown in Fig.~\ref{fig:flowpp}, and the main difference is the dimensionality of the flow. 
Variational dequantization is deployed according to Sec.~\ref{sec:dequant}.
We choose the network architecture for  $q(\zv | \xv)$  to be similar with the variational dequantization network $r(\uv | \xv)$ of Flow++. 
A detailed description of the model architecture is in Appendix~\ref{sec:model-arch}. While we only consider Flow++ for this section due to its impressive density estimation result, our variational data augmentation framework is general and can be combined with future advances of the model architecture. 

The model size is controlled by three main hyper-parameters, (1) the dimensionality of the flow; (2) the hidden layer size $D_H$; and (3) the number of hidden layers $B$ per transformation step. For brevity, we refer to a $32\times 32\times C$-dimensional flow as a $C$-channel flow, where a $3$-channel flow is the baseline Flow++.

The model is trained with an Adam optimizer~\cite{kingma2014adam} with a batch size 64 for 2,000 epochs.
Following~\cite{ho2019flow++}, the learning rate linearly warms up to 0.0012 during the first 2,000 training steps, and exponentially decays at a rate of 0.99999 per step starting from the 50,000-th step until it reaches 0.0003. All the experiments are run on 16 RTX 2080Ti GPUs.
The model quality is measured by bits per dimension (bpd)~\cite{van2016pixel}, where smaller bpd implies higher likelihood and better modeling quality. 
%The dataset is kept with its original 50000 / 10000 split. 
%Without further notice, we report bpd on the test dataset, where the 
The likelihood $P(\xv; \thetav)$ is evaluated with importance sampling as Eq.~(\ref{eqn:importance-sampling}) with $S=4096$ samples for CIFAR-10 and $S=1024$ for ImageNet.

\begin{figure}[t]
\centering
\includegraphics[width=0.8\linewidth]{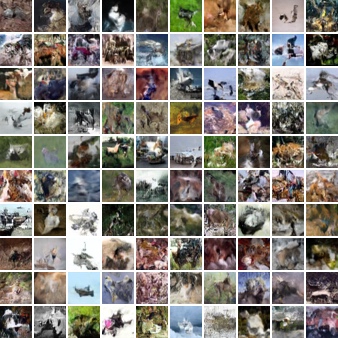}\\
(a) Flow++ (3.08 bpd) \\\vspace{.5em}
\includegraphics[width=0.8\linewidth]{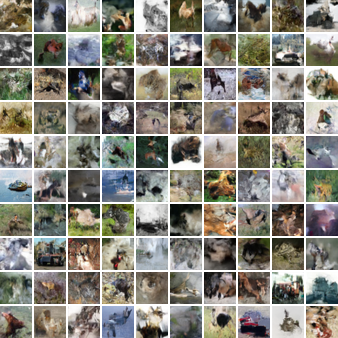}\\
(b) VFlow (2.98 bpd)\\
\caption{Random samples, where (a) is reprinted from~\cite{ho2019flow++}.}
\label{fig:samples}
\end{figure}

\subsection{Improving Existing Models}\label{sec:improving}
We compare a 6-channel VFlow with existing generative flows in Table~\ref{tab:sota}, where the hyperparameters $D_H=96$ and $B=10$ are set identical to Flow++. 
By augmenting the number of channels from 3 to 6, 
For CIFAR-10, VFlow improves the bpd from 3.08 of Flow++ to 2.98. Samples from Flow++ and VFlow  are shown in Fig.~\ref{fig:samples}.

% , where the image quality of VFlow is similar or slightly better than Flow++. Quantitively, the Fr\'echet Inception Distance~\cite{heusel2017gans} improves from 69.1 (Flow++) to 64.9 (VFlow). 
% Therefore, VFlow not only improves the likelihood, but also improves the image quality. We emphasize that this study focuses on whether VFlow can improve existing models, and it is possible to generate more visually appealing images by combining   VFlow with other architectures that have better image quality, such as Residual Flow~\cite{chen2019residual}.

%Comparing with other types of models, VFlow outperforms VAE-IAF~\cite{kingma2016improved}, which is the state-of-the-art VAE. This is reasonable as VFlow generalizes VAE. On the other hand, the autoregressive sparse transformer~\cite{child2019sparsetransformer} achieves a superior 2.80 bpd. However autoregressive models suffer from slow sampling. 

\subsection{Ablation Study under Fixed Parameter Budget}
\label{sec:ablation}
\begin{figure}
\centering
\includegraphics[width=.8\linewidth]{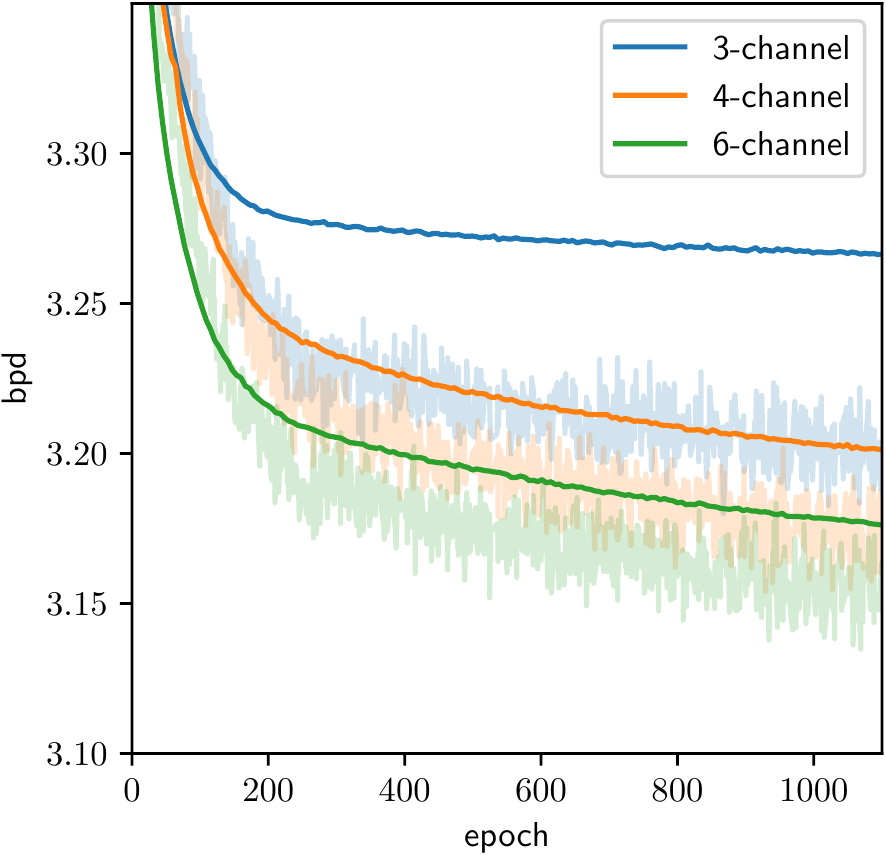}
\caption{Bpd on training (light) and validation (dark) dataset of Flow++ and VFlow under a 4-million parameter budget (not fully converged). Here bpd is only a upper bound because we evaluate it with ELBO as Eq.~(\ref{eqn:full-elbo}) instead of the marginal likelihood. 
\label{fig:curve_expb}}
\end{figure}

To further investigate the impact of the dimensionality, we vary the number of channels under a fixed 4 million parameter budget on the CIFAR-10 dataset. In this set of experiments, we  randomly hold out 10,000 samples from the training set for validation. 
As the dimensionality grows, we reduce the number of hidden layers $B$ to stay within the parameter budget. 
The training curve and final bpd are reported in Fig.~\ref{fig:curve_expb} and Table~\ref{tab:expb} respectively, which clearly show that increased dimensionality is beneficial. Moreover, VFlow achieves better results at all stages of training. 
This supports Remark 1 in Sec.~\ref{sec:connection-to-flows} that VFlow is still better even taken optimization issues into account. For this $D_H=32$ network, a 6-channel VFlow, which has 24 channels on the $16\times 16$ scale, is already sufficient to resolve the bottleneck problem. Going beyond 6 channels has marginal improvement on the model quality, while increasing the number of parameters. 

Interestingly, Fig.~\ref{fig:curve_expb} suggests that besides improved model capacity, the generalization gap of VFlow is also slightly smaller than Flow++. We suspect the additional randomness introduced by $\zv$ acts as an implicit regularization. %Finally, Appendix C provides additional study on the size of the $q(\zv | \xv)$ network. 

\begin{table}[t]
    \centering
    \caption{Impact of dimensionality on the CIFAR-10 dataset.  }
    \vskip 0.15in
    \begin{small}
    \begin{tabular}{lcccc}
    \toprule
    Model & bpd & Parameters & $D_H$ & $B$ \\
    \midrule
        3-channel Flow++ & 3.23 & 4.02M & 32 & 13 \\
        4-channel VFlow & 3.16 & 4.03M & 32 & 11 \\
        6-channel VFlow & \textbf{3.13} & \textbf{4.01M} &  32 & 10 \\
        %\hline
        %PixelCNN~\cite{van2016pixel} & & 65.93 \\
    \bottomrule
    \end{tabular}
    \end{small}
    \label{tab:expb}
\end{table}

\begin{table}
    \centering
    \caption{Parameter efficiency on CIFAR-10.}
    \vskip 0.15in
    \begin{small}
    \begin{tabular}{lcccc}
    \toprule
    Model & bpd & Parameters & $D_H$ & $B$ \\
    \midrule
        3-channel Flow++ & 3.08 & 31.4M & 96 & 10 \\
        6-channel VFlow & \textbf{2.98} & 37.8M & 96 & 10 \\
        6-channel VFlow & 3.03 & 16.5M & 64 & 10\\
        6-channel VFlow & 3.08 & \textbf{11.9M} & 56 & 10\\
    \bottomrule
    \end{tabular}
    \end{small}
    \label{tab:param-efficiency}
\end{table}

\subsection{Parameter Efficiency}\label{sec:exp-param-efficiency}
The last set of experiments aims for more compact models with similar model capacity. As shown by Table~\ref{tab:param-efficiency}, we can reduce the number of parameters of the baseline Flow++ from 31.4 million to 11.9 million, which is a 2.6 times reduction. This reduction of model size is achieved by reducing the hidden layer size $D_H$ from 96 to 56. As we argued in Sec.~\ref{sec:bottleneck-problem}, the excessive number of hidden units does not help much for a network with merely 3 channels. Increasing the dimensionality of the network (\emph{i.e.,} the bottleneck width) is much more efficient than increasing $D_H$. Therefore, VFlow is not only more expressive but also more compact than vanilla low-dimensional flows. We emphasize that the reduction of model size come solely from resolving the bottleneck problem. Even smaller models can be forged by combining with potentially more compact architectures, such as MintNet~\cite{song2019mintnet}.

\section{Conclusions}
We identify the bottleneck problem which limits the capacity of generative flows. To tackle this problem, we propose VFlow, a variational data augmentation framework that pads extra dimensions to the data with learnable variational distributions for the padded data. VFlow is a generalization of vanilla generative flows, and can be combined with existing generative flows to improve their expressiveness and compactness. In our experiments on the CIFAR-10 dataset, VFlow achieves a new state-of-the-art 2.98 bpd, while retaining the 3.08 bpd of vanilla Flow++ with 2.6 times less parameters.

\section*{Acknowledgements}

We thank Chongxuan Li, Yucen Luo, Ziyu Wang, Kun Xu, Fan Bao, Shihong Song and Qian Fu for proofreading. This work was supported by the National Key Research and Development Program of China (No. 2017YFA0700904), NSFC Projects (Nos. 61620106010, U19B2034, U1811461), Beijing NSF Project (No. L172037), Beijing Academy of Artificial Intelligence (BAAI), Tsinghua-Huawei Joint Research Program, Huawei Hisilicon Kirin solution, the MindSpore team, a grant from Tsinghua Institute for Guo Qiang, Tiangong Institute for Intelligent Computing,  and the NVIDIA NVAIL Program with GPU/DGX Acceleration.

\bibliography{vflow}
\bibliographystyle{icml2020}

\clearpage
\newpage
\onecolumn

\appendix
\icmltitle{VFlow: More Expressive Generative Flows with \\ Variational Data Augmentation\\Supplementary Material}

\section{Verification of Assumption A1 and A2}\label{sec:assumptions}
\noindent\textbf{A}1\quad
%For each $\thetav_{D_X}\in \Theta_{D_X}$, there exists $\thetav_{D_X+D_Z}\in \Theta_{D_X+D_Z}$, such that 
%$$\fv_{D_X+D_Z}(\xv, \zv; \thetav_{D_X+D_Z})=\left(\fv_{D_X}(\xv; \thetav_{D_X}), \zv\right).$$
For all $p(\xv; \thetav_{D_X})\in \Pc_{D_X}$ and $D_Z> 0$,
there exists $p(\xv, \zv; \thetav_{D_X+D_Z})\in \Pc_{D_X+D_Z}$, such that for all $\xv$ and $\zv$, 
$$p(\xv, \zv; \thetav_{D_X+D_Z})=p(\xv; \thetav_{D_X})p_{\epsilonv}(\zv).$$

\noindent\textbf{A}2\quad
%There exists $\phiv$ such that $\gv(\epsilonv_q; \xv, \phiv)=\epsilonv_q$. In other words, the flow family has an identity mapping.
For all $D_Z> 0$, there exists $q(\zv | \xv; \phiv)\in \Qc_{D_Z}$, such that for all $\xv$ and $\zv$,
$$q(\zv | \xv; \phiv)=p_{\epsilonv}(\zv).$$ 

\vskip 0.75em

\newcommand{\cc}[2]{\begin{bmatrix}{#1} & {#2}\end{bmatrix}}
\newcommand{\xz}{\cc{\xv}{\zv}}

Let $\xv, \zv$ be row vectors, and $\xz$ be the horizontal concatenation of $\xv$ and $\zv$. We first show that the following conditions are sufficient for Assumption A1 and A2. 

\noindent\textbf{B}1\quad
For all $\thetav_{D_X}\in \Theta_{D_X}$ and $D_Z> 0$,
there exists $\thetav_{D_X+D_Z}\in \Theta_{D_X+D_Z}$, such that for all $l$, $\xv$ and $\zv$, 
$$\fv_l(\xz; \thetav_{D_X+D_Z})=\cc{\fv_l(\xv; \thetav_{D_X})}{\zv}.$$

\noindent\textbf{B}2\quad
For all $D_Z> 0$, there exists $\phiv\in \Phi_{D_Z}$, such that for all $l$, $\xv$ and $\zv$, 
$$\gv_l(\epsilonv_q; \xv, \phiv)=\epsilonv_q.$$ 

\vskip 0.75em

\begin{proof}
Under condition B1, 
\begin{align*}
&\fv(\xz)\\
=&\fv_1(\dots(\fv_L(\xz)))\\
=&\fv_1(\dots(\fv_{L-1}( \cc{\fv_L(\xv)}{\zv} )))\\
=&\fv_1(\dots(\fv_{L-2}( \cc{\fv_{L-1}(\fv_L(\xv))}{\zv} )))=\dots\\
=&\cc{\fv_1(\dots(\fv_L(\xv)))}{\zv}.    
\end{align*}
Then, 
\begin{align*}
p(\xv, \zv; \thetav_{D_X+D_Z})&=p_{\epsilonv}(\fv(\xz; \thetav_{D_X+D_Z}))\abs{\frac{\partial \fv(\xz; \thetav_{D_X+D_Z})}{\partial \xz}} \\
&= p_{\epsilonv}(\cc{\fv(\xv; \thetav_{D_X})}{\zv})\abs{\frac{\partial \cc{\fv(\xv; \thetav_{D_X})}{\zv}}{\partial \xz}}\\
&= p_{\epsilonv}(\fv(\xv; \thetav_{D_X}))p_{\epsilonv}(\zv)\abs{\begin{bmatrix}
\frac{\partial \fv(\xv; \thetav_{D_X})}{\partial \xv} & \mathbf{0}\\
\mathbf{0} & \Iv 
\end{bmatrix}}\\
&= p_{\epsilonv}(\fv(\xv; \thetav_{D_X}))\abs{\frac{\partial \fv(\xv; \thetav_{D_X})}{\partial \xv}}p_{\epsilonv}(\zv) \\
&= p(\xv; \thetav_{D_X})p_{\epsilonv}(\zv).
\end{align*}
Similarly, under condition B2,
$$\gv(\epsilonv_q; \xv, \phiv)=\gv_1(\dots(\gv_L(\epsilonv_q)))=\epsilonv_q.$$
So
$$q(\zv; \xv, \phiv)=q(\gv(\epsilonv_q; \xv, \phiv) |\xv; \phiv)=p_{\epsilonv}(\epsilonv_q)/\abs{\frac{\partial \zv}{\partial \epsilonv_q}}=p_{\epsilonv}(\epsilonv_q)/\abs{\Iv}=p_{\epsilonv}(\epsilonv_q). $$
\end{proof}

Therefore, we only need to verify condition B1 and B2 separately for each transformation step . For Glow~\cite{glow} and Residual Flow~\cite{chen2019residual}, the transformations to verify includes affine coupling layer~\cite{realnvp}, invertible 1$\times$1 convolution~\cite{glow},  and invertible residual blocks~\cite{behrmann2019invertible}. In this section we only verify condition B1 and B2 for fully-connected transformations, but they readily generalize to convolutional transformations. 

\subsection{Invertible Residual Blocks}
\begin{figure}
    \centering
    \includegraphics[width=.5\linewidth]{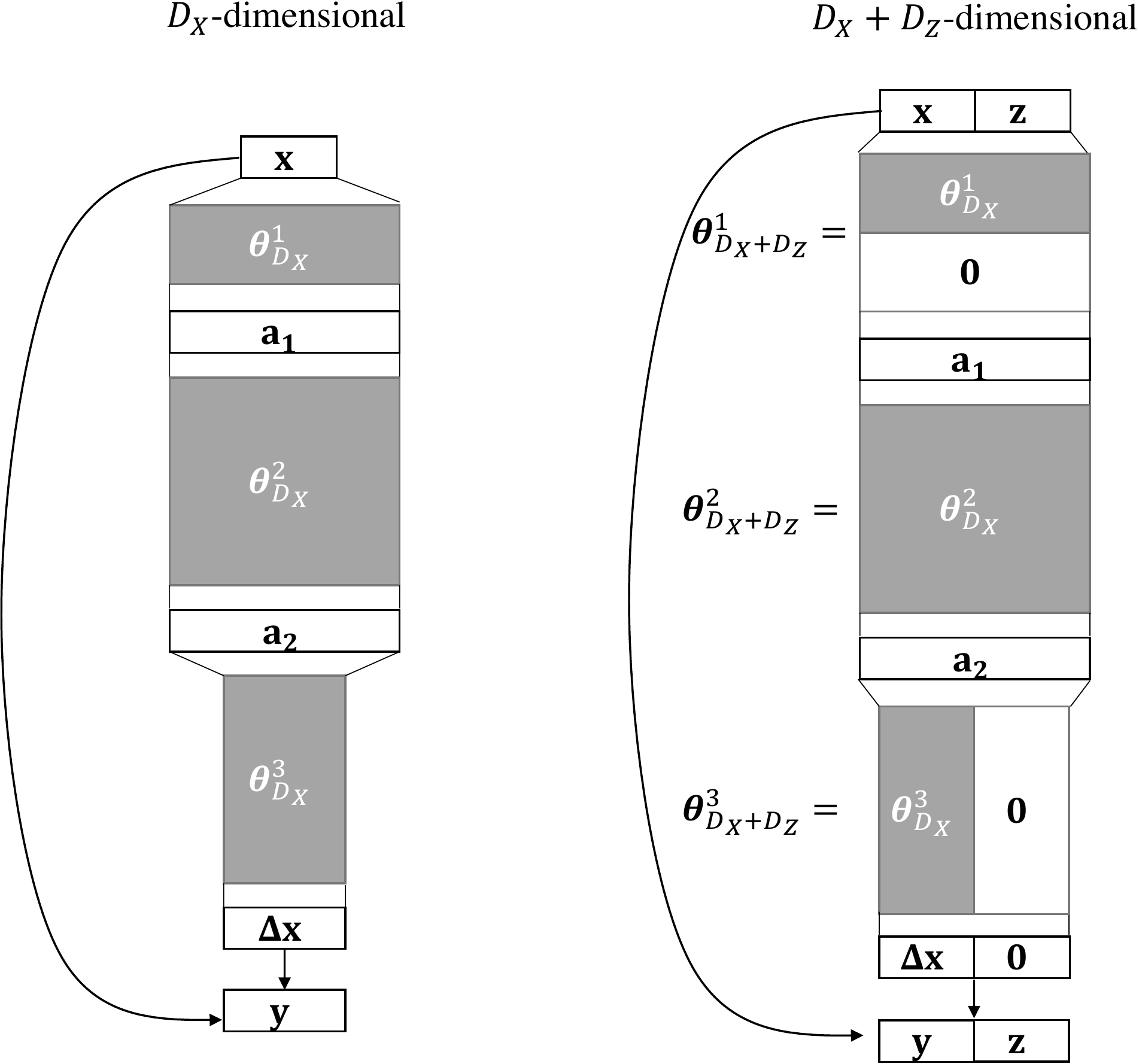}
    \caption{Constructing $\fv_l(\xz; \thetav_{D_X+D_Z})$ based on $\fv_l(\xv; \thetav_{D_X})$.}
    \label{fig:appendix-a}
\end{figure}
An invertible residual block~\cite{behrmann2019invertible} $\fv_l(\xv; \thetav_{D_X})$ for $D_X$-dimensional input $\xv$ is defined as
\begin{align*}
\av_1 = \xv \thetav_{D_X}^{(1)},\quad 
\av_2 = \nv(\av_1; \thetav_{D_X}^{(2)}),\quad 
\Delta_{\xv} = \av_2 \thetav_{D_X}^{(3)},\quad 
\fv_l(\xv; \thetav_{D_X})=\yv=\xv+\Delta_{\xv},
\end{align*}
where we explicitly write the first and last linear layer, and leave all the internal hidden layers as $\nv(\av_1; \thetav_{D_X}^{(2)})$. We construct a $D_X+D_Z$-dimensional invertible residual block $\fv_l(\xz; \thetav_{D_X+D_Z})$ as
\begin{align*}
&\av_1 = \xz \thetav_{D_X+D_Z}^{(1)},\quad 
\av_2 = \nv(\av_1; \thetav_{D_X+D_Z}^{(2)}),\quad 
\cc{\Delta_{\xv}}{\mathbf{0}} = \av_2 \thetav_{D_X+D_Z}^{(3)},\\
&\fv_l(\xz; \thetav_{D_X+D_Z})=\cc{\xv+\Delta_{\xv}}{\zv+\mathbf{0}}=\cc{\fv_l(\xv; \thetav_{D_X})}{\zv},
\end{align*}
satisfying condition B1, where
\begin{align*}
\thetav_{D_X+D_Z}^{(1)}= \begin{bmatrix}\thetav_{D_X}^{(1)}\\ \mathbf{0}\end{bmatrix}, \quad \thetav_{D_X+D_Z}^{(2)}= \thetav_{D_X}^{(2)},\quad \thetav_{D_X+D_Z}^{(3)}=\cc{\thetav_{D_X}^{(3)}}{\mathbf{0}}.
\end{align*}
This construction is demonstrated by Fig.~\ref{fig:appendix-a}. Intuitively, due to the residual structure, we only need to output $\mathbf{0}$ for all the $\zv$ dimensions. Similarly, condition B2 can be satisfied by taking $\thetav_{D_Z}^{(3)}=\mathbf{0}$, so that all the residuals are zero and the network outputs identity.

\subsection{Affine Coupling Layer}
An affine coupling layer~\cite{realnvp} $\fv_l(\xv; \thetav_{D_X})$ for $D_X$-dimensional input $\xv$ is defined as
\begin{align*}
\xv_1, \xv_2 = \splitop(\xv),\quad \yv_1=\xv_1,\quad \yv_2=\muv(\xv_1; \thetav_{D_X})+\exp(\sv(\xv_1; \thetav_{D_X}))\circ \xv_2,\quad 
\fv_l(\xv; \thetav_{D_X})=\concat(\yv_1, \yv_2).
\end{align*}
The case of affine coupling layer is almost identical to the invertible residual block, because both transformations have residual structures. This can be seen by noticing when $\muv(\xv_1; \thetav_{D_X})=\sv(\xv_1; \thetav_{D_X})=\mathbf{0}$, $\fv_l(\xv; \thetav_{D_X})=\xv$. 
We explicitly write out the first and last linear layers of $\muv(\cdot)$ and $\sv(\cdot)$:
\begin{align*}
&\xv_1, \xv_2 = \splitop(\xv),\quad \yv_1=\xv_1,\\
&\av_1=\xv_1  \thetav_{D_X}^{(a1)},\quad \av_2=\muv^\prime(\av_1; \thetav_{D_X}^{(a2)}), \quad \av_3=\av_2\thetav_{D_X}^{(a3)},\\
&\bv_1=\xv_1  \thetav_{D_X}^{(b1)},\quad \bv_2=\sv^\prime(\bv_1; \thetav_{D_X}^{(b2)}), \quad \bv_3=\bv_2\thetav_{D_X}^{(b3)},\\
&\yv_2=\av_3+\bv_3\circ \xv_2,\quad \fv_l(\xv; \thetav_{D_X})=\concat(\yv_1, \yv_2).
\end{align*}
A $D_X+D_Z$-dimensional affine coupling layer has the form
\begin{align*}
&\cc{\xv_1}{\zv_1}, \cc{\xv_2}{\zv_2} = \splitop(\xz),\quad \yv_1=\xv_1,\\
&\av_1=\cc{\xv_1}{\zv_1}  \thetav_{D_X+D_Z}^{(a1)},\quad \av_2=\muv^\prime(\av_1; \thetav_{D_X+D_Z}^{(a2)}), \quad \cc{\av_3}{\uv_3}=\av_2\thetav_{D_X+D_Z}^{(a3)},\\
&\bv_1=\cc{\xv_1}{\zv_1}  \thetav_{D_X+D_Z}^{(b1)},\quad \bv_2=\sv^\prime(\bv_1; \thetav_{D_X+D_Z}^{(b2)}), \quad \cc{\bv_3}{\wv_3}=\bv_2\thetav_{D_X+D_Z}^{(b3)},\\
&\yv_2=\cc{\av_3+\bv_3\circ \xv_2}{\uv_3+\wv_3\circ \zv_2},\quad \fv_l(\xz; \thetav_{D_X+D_Z})=\concat(\cc{\yv_1}{\zv_1}, \yv_2),
\end{align*}
We want the networks $\muv(\cdot; \thetav_{D_X+D_Z})$ and $\sv(\cdot; \thetav_{D_X+D_Z})$ to ignore the $\zv_1$ part from the input, and output zero for $\uv_3$, $\wv_3$, so that 
\begin{align*}
&\cc{\av3}{\uv_3} = \cc{\muv({\xv_1}; \thetav_{D_X})}{\mathbf{0}}, \quad \cc{\bv3}{\wv_3} = \cc{\sv({\xv_1}; \thetav_{D_X})}{\mathbf{0}}\\
&\yv_2=\cc{\muv(\xv_1; \thetav_{D_X})+\exp(\sv(\xv_1; \thetav_{D_X}))\circ \xv_2}{\zv_2},\quad \fv_l(\xz; \thetav_{D_X+D_Z})=\cc{\fv_l(\xv; \thetav_{D_X})}{\zv},
\end{align*}
so condition B1 is satisfied. We can easily achieve this by setting
\begin{align*}
&\thetav_{D_X+D_Z}^{(a1)}
=\begin{bmatrix}\thetav_{D_X}^{(a1)}\\ \mathbf{0}\end{bmatrix}, \quad \thetav_{D_X+D_Z}^{(a2)}= \thetav_{D_X}^{(a2)},\quad \thetav_{D_X+D_Z}^{(a3)}=\cc{\thetav_{D_X}^{(a3)}}{\mathbf{0}}\\
&\thetav_{D_X+D_Z}^{(b1)}
=\begin{bmatrix}\thetav_{D_X}^{(b1)}\\ \mathbf{0}\end{bmatrix}, \quad \thetav_{D_X+D_Z}^{(b2)}= \thetav_{D_X}^{(b2)},\quad \thetav_{D_X+D_Z}^{(b3)}=\cc{\thetav_{D_X}^{(b3)}}{\mathbf{0}}.
\end{align*}
Similarly, condition B2 can be satistied by setting $\thetav_{D_Z}^{(a3)}=\thetav_{D_Z}^{(b3)}=\mathbf{0}$.

\subsection{Invertible 1$\times$1 Convolution}
For fully-connected cases, invertible 1$\times$1 convolution~\cite{glow} degenerates to a regular matrix multiplication
\begin{align*}
\fv_l(\xv; \thetav_{D_X})=\xv\thetav_{D_X},
\end{align*}
where $\thetav_{D_X}$ is a non-singular matrix. We construct $\fv_l(\xv; \thetav_{D_X+D_Z})$ such that 
\begin{align*}
\thetav_{D_X+D_Z}=\begin{bmatrix}
\thetav_{D_X} & \mathbf{0}\\
\mathbf{0} & \Iv
\end{bmatrix}.
\end{align*}
Clearly, $\thetav_{D_X+D_Z}$ is also non-singular, and $\fv_l(\xz; \thetav_{D_X+D_Z})=\cc{\fv_l(\xv; \thetav_{D_X})}{\zv}$. On the other hand, condition B2 can be satisfied by setting $\thetav_{D_Z}=\Iv$.

\begin{table}[t]
    \centering
    \caption{Model architecture for improving existing models experiment and parameter efficiency experiment.}
    \begin{tabular}{c|c|c|c|c}
    \toprule
        Model & 3-channel Flow++ & 6-channel VFlow & 6-channel VFlow & 6-channel VFlow\\
    \hline
        Parameters & 31.4M & 37.8M & 16.5M & 11.9M\\
    \hline
        bpd & 3.08 & 2.98 & 3.03 & 3.08\\
    \bottomrule
         \multicolumn{5}{c}{Architecture for $p(\xv,\zv)$: direction $(\xv, \zv)\rightarrow\epsilonv$ }\\
    \toprule
        \multirow{2}{*}{$32\times32$} & \multirow{2}{*}{$f_{\mathrm{checker}}(10,96,32)$ $\times 4$} & $f_{\mathrm{checker}}(10,96,32)$ $\times 2$ & $f_{\mathrm{checker}}(10,64,16)$ $\times 2$ & $f_{\mathrm{checker}}(10,56,10)$ $\times 2$\\
        & & $f_{\mathrm{channel}}(10,96,32)$ $\times 2$ & $f_{\mathrm{channel}}(10,64,16)$ $\times 2$ & $f_{\mathrm{channel}}(10,56,10)$ $\times 2$\\
    \hline
        & SpaceToDepth & SpaceToDepth & SpaceToDepth & SpaceToDepth\\
    \hline
        \multirow{2}{*}{$16\times16$} & $f_{\mathrm{channel}}(10,96,32)$ $\times 2$ & $f_{\mathrm{checker}}(10,96,32)$ $\times 2$ & $f_{\mathrm{checker}}(10,64,16)$ $\times 2$ & $f_{\mathrm{checker}}(10,56,10)$ $\times 2$\\
        & $f_{\mathrm{checker}}(10,96,32)$ $\times 3$ & $f_{\mathrm{channel}}(10,96,32)$ $\times 3$ & $f_{\mathrm{channel}}(10,64,16)$ $\times 3$ & $f_{\mathrm{channel}}(10,56,10)$ $\times 3$\\
    \bottomrule
        \multicolumn{5}{c}{Architecture for $q(\zv|\xv)$: direction   $\epsilonv_q\rightarrow\zv$}\\
    \toprule
        \multirow{2}{*}{$32\times32$} & \multirow{2}{*}{N/A} & $g_{\mathrm{checker}}(3,96,32)$ $\times 4$ & $g_{\mathrm{checker}}(3,64,16)$ $\times 4$ & $g_{\mathrm{checker}}(3,56,10)$ $\times 4$ \\
        & & Sigmoid & Sigmoid & Sigmoid\\
    \hline
        \multicolumn{5}{c}{Architecture for $r(\uv|\xv)$: direction   $\epsilonv_r\rightarrow\uv$}\\
    \hline
        \multirow{2}{*}{$32\times32$} & $g_{\mathrm{checker}}(2,96,32)$ $\times 4$ & $g_{\mathrm{checker}}(2,96,32)$ $\times 4$ & $g_{\mathrm{checker}}(2,64,16)$ $\times 4$ & $f_{\mathrm{checker}}(2,56,10)$ $\times 4$ \\
        & Sigmoid & Sigmoid & Sigmoid & Sigmoid\\
    \bottomrule
    \end{tabular}
    \label{tab:exp_ac_arch}
\end{table}

\begin{table}[t]
    \centering
    \caption{Model architecture for ablation experiment under fixed parameter budget.}
    \begin{tabular}{c|c|c|c}
    \toprule
        Model & 3-channel Flow++ & 4-channel VFlow & 6-channel VFlow\\
    \hline
        Parameters & 4.02M & 4.03M & 4.01M\\
    \hline
        bpd & 3.21 & 3.15 & 3.12\\
    \bottomrule
        \multicolumn{4}{c}{Architecture for $p(\xv,\zv)$: direction $(\xv, \zv)\rightarrow\epsilonv$}\\
    \toprule
        \multirow{3}{*}{$32\times32$} & \multirow{3}{*}{$f_{\mathrm{checker}}(13,32,4)$ $\times 4$} & $f_{\mathrm{affine}}(3,32)$ $\times 1$ & $f_{\mathrm{affine}}(3,32)$ $\times 1$\\
        & & $f_{\mathrm{checker}}(11,32,4)$ $\times 2$ & $f_{\mathrm{checker}}(10,32,4)$ $\times 2$\\
        & & $f_{\mathrm{channel}}(11,32,4)$ $\times 2$ & $f_{\mathrm{channel}}(10,32,4)$ $\times 2$\\
    \hline
        & SpaceToDepth & SpaceToDepth & SpaceToDepth\\
    \hline
        \multirow{2}{*}{$16\times16$} & $f_{\mathrm{channel}}(13,32,4)$ $\times 2$ & $f_{\mathrm{checker}}(11,32,4)$ $\times 2$ & $f_{\mathrm{checker}}(10,32,4)$ $\times 2$\\
        & $f_{\mathrm{checker}}(13,32,4)$ $\times 3$ & $f_{\mathrm{channel}}(11,32,4)$ $\times 3$ & $f_{\mathrm{channel}}(10,32,4)$ $\times 3$\\
    \bottomrule
        \multicolumn{4}{c}{Architecture for $q(\zv|\xv)$: direction   $\epsilonv_q\rightarrow\zv$}\\
    \toprule
        \multirow{2}{*}{$32\times32$} & \multirow{2}{*}{N/A} & $g_{\mathrm{checker}}(3,32,4)$ $\times 4$ & $g_{\mathrm{checker}}(3,32,4)$ $\times 4$\\
        & & Sigmoid & Sigmoid\\
    \hline
        \multicolumn{4}{c}{Architecture for $r(\uv|\xv)$: direction   $\epsilonv_r\rightarrow\uv$}\\
    \hline
        \multirow{2}{*}{$32\times32$} & $f_{\mathrm{checker}}(2,32,4)$ $\times 4$ & $f_{\mathrm{checker}}(2,32,4)$ $\times 4$ & $f_{\mathrm{checker}}(2,32,4)$ $\times 4$\\
        & Sigmoid & Sigmoid & Sigmoid\\
    \bottomrule
    \end{tabular}
    \label{tab:exp_b_arch}
\end{table}

\section{Model Architecture}\label{sec:model-arch}

% \begin{itemize}
%     \item Define some layers
%     \item For each experiment
%     \begin{itemize}
%         \item Define the combination of layers (a table)
%         \item Discussion (a total of one paragraph)
%         \item Other minor changes
% \end{itemize}
% \end{itemize}

Our model architecture is directly taken from Flow++~\cite{ho2019flow++}, with some minor changes to make best use of the increased dimensionality. %We also incorporate a fix of the gradient explosion problem of Flow++. 

Flow++ has three types of invertible transformation steps, activation normalization \textbf{ActNorm}~\cite{glow}, invertible 1$\times$1 convolution \textbf{Pointwise}~\cite{glow} and mixture-of-logistic attention coupling layer \textbf{MixLogisticAttnCoupling}~\cite{ho2019flow++}. 
Each coupling layers is controlled by the number of convolution-attention hidden layers $B$, number of filters $D_H$, and number of logistic mixture components $K$, as mentioned in Sec.~\ref{sec:cifar10}. There are two types of input splits for coupling layer, where \textbf{ChannelSplit} partitions input by channel, and \textbf{CheckerboardSplit} partitions input by space. Squeezing operation \textbf{SpaceToDepth}~\cite{realnvp}  is   adopted for multiscale modeling. Conditional distributions, including augmented data distribution $q(\zv | \xv + \uv)$ and dequantization distribution $r(\uv | \xv)$, are implemented by adding a transformed version of $\xv$ to the input of every coupling layer. 
Further denoting \textbf{TupleFlip} as flipping the two split inputs,
\textbf{Inverse($\cdot$)} as the inverse transformation,  and  \textbf{MixLogisticCoupling} as MixLogisticAttnCoupling without attention, Flow++ consists  the following building blocks:
\begin{align*}
    f_{\mathrm{checker}}(B, D_H, K)=& \mbox{CheckerboardSplit}\longrightarrow\mbox{ActNorm}\longrightarrow\mbox{Pointwise}\longrightarrow \mbox{MixLogisticAttnCoupling}(B, D_H, K)\\
    &\longrightarrow\mbox{TupleFlip}\longrightarrow\mbox{Inverse(CheckerboardSplit)} \\
    \\
    f_{\mathrm{channel}}(B, D_H, K)=&
    \mbox{ChannelSplit}\longrightarrow\mbox{ActNorm}\longrightarrow\mbox{Pointwise}\longrightarrow\mbox{MixLogisticAttnCoupling}(B, D_H, K)\\
    &\longrightarrow\mbox{TupleFlip}\longrightarrow\mbox{Inverse(ChannelSplit)} \\
    \\
    g_{\mathrm{checker}}(B, D_H, K)=& \mbox{CheckerboardSplit}\longrightarrow\mbox{ActNorm}\longrightarrow\mbox{Pointwise}\longrightarrow \mbox{MixLogisticCoupling}(B, D_H, K)\\
    &\longrightarrow\mbox{TupleFlip}\longrightarrow\mbox{Inverse(CheckerboardSplit)} \\
    \\
    g_{\mathrm{channel}}(B, D_H, K)=&
    \mbox{ChannelSplit}\longrightarrow\mbox{ActNorm}\longrightarrow\mbox{Pointwise}\longrightarrow\mbox{MixLogisticCoupling}(B, D_H, K)\\
    &\longrightarrow\mbox{TupleFlip}\longrightarrow\mbox{Inverse(ChannelSplit)} \\
\end{align*}

We show the model architectures used in Sec.~\ref{sec:improving} and  Sec.~\ref{sec:exp-param-efficiency} in Table~\ref{tab:exp_ac_arch}, where the architecture of VFlow is almost identical with the baseline Flow++, except we use both $f_{\mathrm{checker}}$ and $f_{\mathrm{channel}}$ for the $32\times 32$ scale, while Flow++ uses only $f_{\mathrm{checker}}$. Flow++ cannot use $f_{\mathrm{channel}}$ for the $32\times 32$ scale because there are odd number (3) of channels. The model architectures under 4-million-parameter budget used in Sec.~\ref{sec:ablation}  are listed in Table~\ref{tab:exp_b_arch}.  In this experiment, we use a special affine coupling layer to mix $\zv$ and $\xv$ forcibly:
\begin{align*}
&\yv_1 = \zv, \quad \yv_2 = \muv(\zv)+\exp(\sv(\zv))\circ \xv,\quad f_{\mathrm{affine}}=\mathrm{concat}(\yv_1,\yv_2)
\end{align*}
where $\muv$ and $\sv$ are $\mathbb{R}^{D_Z}\rightarrow \mathbb{R}^{D_X}$ functions. We empirically find that adding this special affine coupling layer accelerates the convergence for small networks. The building block with this affine coupling layer with $B$ hidden layers and $D_H$ hidden units is denoted as $f_{\mathrm{affine}}(B, D_H)$ in Table~\ref{tab:exp_b_arch}.

\noindent\textbf{Fixing Gradient Explosion}

In our experiments, we find that the implementation of mixture-of-logistic attention coupling layer~\cite{ho2019flow++} sometimes produces huge gradients, leading the training to diverge. 
To see this, note that the mixture-of-logistic attention coupling layer for a given input $\xv=(\xv_1,\xv_2)$ and the output $\yv=(\yv_1,\yv_2)$ is defined by:
\begin{align*}
    &\mathrm{MixLogCDF}(x;\piv,\muv,\sv)=\sum_{i=1}^K \pi_i\sigma((x-\mu_i)\cdot \exp(-s_i)),\quad \mathrm{where}\quad \sum_{i=1}^K\pi_i=1\\
    &\yv_1=\xv_1,\quad \yv_2=\sigma^{-1}\left(\mathrm{MixLogCDF}(\xv_2;\piv_\theta(\xv_1),\mu_\theta(\xv_1),\sv_\theta(\xv_1))\right)\circ \exp(\av_\theta(\xv_1))+\bv_\theta(\xv_1),
\end{align*}
where $\sigma(x)=\frac{1}{1+e^{-x}}$ is the sigmoid function. However, the inverse sigmoid may cause gradient explosion. For example, if $x=1-10^{-N}$, then $\left(\sigma^{-1}(x)\right)^\prime = \frac{1}{x(1-x)}\approx 10^N$. 
If for each component, $x-\mu_i$ is large and $s_i$ is small, then $(x-\mu_i)\cdot \exp(-s_i)$ is large, and $\mathrm{MixLogCDF}(\xv_2;\piv_\theta(\xv_1),\mu_\theta(\xv_1),\sv_\theta(\xv_1)$ will be close to 1, leading to gradient explosion of the inverse sigmoid function. For example, if $\pi_i=1$, $x-\mu_i=4$ and $s_i=-1$, we have $\mathrm{MixLogCDF}=\sigma((x-\mu_i)\cdot \exp(-s_i))\approx 1-2\cdot 10^{-5}$ and then the gradient can be very large. We fix this issue by scaling the input of the inverse sigmoid function to $[0.05, 0.95]$:
\begin{align*}
    \yv_2=\sigma^{-1}(0.05 + 0.9*\mathrm{MixLogCDF}(\xv_2;\piv_\theta(\xv_1),\muv_\theta(\xv_1),\sv_\theta(\xv_1)))\circ \exp(\av_\theta(\xv_1))+\bv_\theta(\xv_1).
\end{align*}

\section{Extra Experiments}
We further study whether it is better to put more parameters on $p(\xv, \zv)$ or $q(\zv | \xv)$. Under a fixed 4 million total parameter budget, we vary the parameter allocation between $p(\xv, \zv)$ or $q(\zv | \xv)$, and list the corresponding result and model architecture in Table~\ref{tab:exp_d_p_and_q}. The result implies that it is better to put most parameters on $p(\xv, \zv)$, supporting our claim in Sec.~\ref{sec:related-works} that the variational distribution of VFlow is not necessarily as complicated as those in VAEs. 

\begin{table}[H]
    \centering
        \caption{Parameter allocation between $p(\xv, \zv)$ and $q(\zv | \xv)$.}
    \begin{tabular}{c|c|c|c}
    \toprule
         & 6-channel VFlow & 6-channel VFlow & 6-channel VFlow\\
    \hline
        Total parameters & 4.00M & 4.05M & 4.01M\\
    \hline
        $q(\zv|\xv)$ parameters & 0.83M & 0.64M & 0.36M\\
    \hline
        bpd & 3.14 & 3.13 & 3.12\\
    \bottomrule
        \multicolumn{4}{c}{Architecture for $p(\xv,\zv)$: direction $(\xv, \zv)\rightarrow\epsilonv$}\\
    \toprule
        \multirow{3}{*}{$32\times32$} & $f_{\mathrm{affine}}(3,32)$ $\times 1$ & $f_{\mathrm{affine}}(3,32)$ $\times 1$ & $f_{\mathrm{affine}}(3,32)$ $\times 1$\\
        & $f_{\mathrm{checker}}(8,32,4)$ $\times 2$ & $f_{\mathrm{checker}}(9,32,4)$ $\times 2$ & $f_{\mathrm{checker}}(10,32,4)$ $\times 2$\\
        & $f_{\mathrm{channel}}(8,32,4)$ $\times 2$ & $f_{\mathrm{channel}}(9,32,4)$ $\times 2$ & $f_{\mathrm{channel}}(10,32,4)$ $\times 2$\\
    \hline
        & SpaceToDepth & SpaceToDepth & SpaceToDepth\\
    \hline
        \multirow{2}{*}{$16\times16$} & $f_{\mathrm{checker}}(8,32,4)$ $\times 2$ & $f_{\mathrm{checker}}(9,32,4)$ $\times 2$ & $f_{\mathrm{checker}}(10,32,4)$ $\times 2$\\
        & $f_{\mathrm{channel}}(8,32,4)$ $\times 3$ & $f_{\mathrm{channel}}(9,32,4)$ $\times 3$ & $f_{\mathrm{channel}}(10,32,4)$ $\times 3$\\
    \bottomrule
        \multicolumn{4}{c}{Architecture for $q(\zv|\xv)$: direction   $\epsilonv_q\rightarrow\zv$}\\
    \toprule
        \multirow{2}{*}{$32\times32$} & $g_{\mathrm{checker}}(8,32,4)$ $\times 4$ & $g_{\mathrm{checker}}(6,32,4)$ $\times 4$ & $g_{\mathrm{checker}}(3,32,4)$ $\times 4$\\
        & Sigmoid & Sigmoid & Sigmoid\\
    \bottomrule
        \multicolumn{4}{c}{Architecture for $r(\uv|\xv)$: direction   $\epsilonv_r\rightarrow\uv$}\\
    \toprule
        \multirow{2}{*}{$32\times32$} & $f_{\mathrm{checker}}(2,32,4)$ $\times 4$ & $f_{\mathrm{checker}}(2,32,4)$ $\times 4$ & $f_{\mathrm{checker}}(2,32,4)$ $\times 4$\\
        & Sigmoid & Sigmoid & Sigmoid\\
    \bottomrule
    \end{tabular}
    \label{tab:exp_d_p_and_q}
\end{table}

\end{document}